\documentclass[letterpaper]{article} %
\usepackage[preprint]{aaai2027}  %
\usepackage[hyphens]{url}  %
\usepackage{graphicx} %
\usepackage{natbib}  %
\usepackage{caption} %
\usepackage{subcaption} %
\usepackage{algorithm}
\usepackage{algorithmic}
\usepackage{amsmath}
\usepackage{amssymb}
\usepackage{booktabs}
\usepackage{xcolor}
\usepackage{colortbl}
\usepackage{multirow}
\usepackage{tabularx}
\usepackage{dblfloatfix} %
\definecolor{mycol1}{HTML}{E9F2FF}
\definecolor{mycol2}{HTML}{DED6D2}
\definecolor{accentcol}{HTML}{F3E8FF}
\definecolor{rvcol}{HTML}{1D4ED8} %
\definecolor{racol}{HTML}{B91C1C} %

\title{Deferred Audio Pruning with Local Audio--Visual Dynamics for Omni-LLMs}

\author{
Kyeongyoon Lee\textsuperscript{\rm 1},
Hongyeob Kim\textsuperscript{\rm 1},
Youngeun Kim\textsuperscript{\rm 2},
Sungeun Hong\textsuperscript{\rm 1}
}
\affiliations{
\textsuperscript{\rm 1}Sungkyunkwan University\\
\textsuperscript{\rm 2}AWS AI Labs
}
\newcommand{\ours}{\texttt{\textbf{A-PACK }}}

\begin{document}

\maketitle

\begin{abstract}
Omni-modal LLMs jointly process audio, video, and text, but long multimodal sequences incur substantial prefill and KV-cache costs. Existing omni-modal compression methods primarily focus on pre-LLM token reduction, leaving modality-specific compression across the LLM boundary underexplored. We propose A-PACK, a two-stage framework that defers audio pruning until query-conditioned multimodal interactions emerge. Our analysis shows that audio exhibits higher task-relevant information density and representational diversity per token than video. We further find that local audio--visual dynamics provide a more effective cue for visual selection than token-wise matching. We therefore preserve audio and compress video with local dynamics before the LLM, then progressively prune low-relevance audio and visual tokens and their KV-cache entries inside the LLM. Across four benchmarks on Qwen2.5-Omni-7B/3B, A-PACK achieves the strongest average performance among the evaluated prior methods while reducing prefill FLOPs by up to 78\% and improving decoding throughput by up to 2.21$\times$.
\end{abstract}

\section{Introduction}

\begin{figure}[ht]
    \centering
    \begin{subfigure}[b]{0.99\columnwidth}
        \centering
        \includegraphics[width=\linewidth]{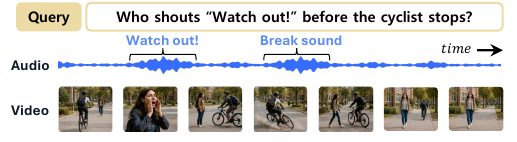}
        \caption{Example query with audio and video frames}
        \label{fig:teasera}
    \end{subfigure}

    \begin{subfigure}[b]{0.99\columnwidth}
        \centering
        \includegraphics[width=\linewidth]{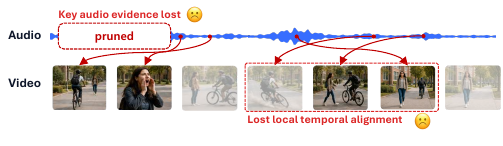}
        \caption{Early pruning and pointwise matching}
        \label{fig:teaserb}
    \end{subfigure}

    \begin{subfigure}[b]{0.99\columnwidth}
        \centering
        \includegraphics[width=\linewidth]{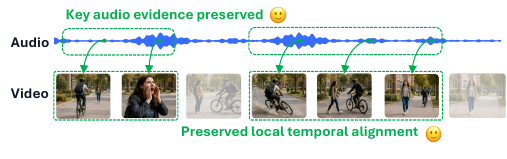}
        \caption{Proposed: deferred audio pruning with local dynamics}
        \label{fig:teaserc}
    \end{subfigure}

\caption{
Motivation of our method.
Early audio pruning may discard compact but important evidence, and pointwise audio--visual matching may associate temporally distant events.
We preserve audio before the LLM and guide visual compression with local audio--visual dynamics. We prune audio later using query-conditioned relevance.
}
    \label{fig:teaser}
\end{figure}

Video understanding has progressed from general vision--language models~\citep{llavaonevision} to architectures that capture richer spatiotemporal information~\citep{videollama2,videochatflash,longvila,longvu}. More recently, omni-modal LLMs have jointly processed audio, video, and text~\citep{qwen25omni,ola,baichuanomni}, enabling tasks that require both listening and watching. Their long audio--visual sequences, however, substantially increase prefill computation and KV-cache memory~\citep{longvu}.

Omni-modal token compression methods make their modality-selection decisions before the LLM, using cross-modal cues~\citep{omnizip}, audio-conditioned recoverability~\citep{contextguard}, query-adaptive budgets~\citep{omniselect}, or asymmetric modality selection~\citep{omnisift}. Importantly, this leaves modality-specific compression across the LLM boundary largely unexplored. Before audio, video, and the query are jointly contextualized within the LLM, they remain represented by modality-specific features. Relevance estimates are therefore incomplete. Early pruning may discard evidence that becomes important during multimodal reasoning, especially brief but informative audio events. In addition, token-wise audio--visual matching~\citep{omnizip,localcorr} can overlook temporal locality and connect semantically similar but non-co-occurring events.

\begin{figure}[t]
  \centering
  \begin{subfigure}[t]{0.50\columnwidth}
    \centering
    \includegraphics[width=\linewidth]{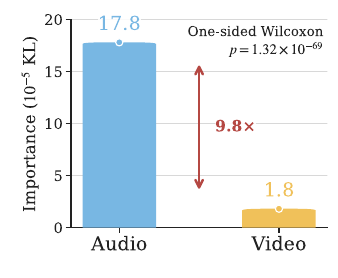}
    \caption{Per-token importance}
  \end{subfigure}\hfill
  \begin{subfigure}[t]{0.50\columnwidth}
    \centering
    \includegraphics[width=\linewidth]{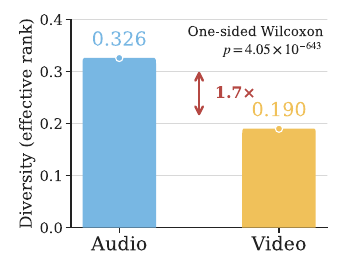}
    \caption{Per-token diversity}
  \end{subfigure}
\caption{Audio tokens show significantly higher per-token importance and diversity than video tokens. Higher occlusion KL indicates greater impact on predictions when removed, while higher effective rank indicates less redundancy.}
  \label{fig:density}
  \end{figure}

Although existing methods vary in token scoring and modality-budget allocation, modality-specific decisions are largely made before the LLM. Our key insight is that the criterion for token relevance changes across the LLM boundary, from modality-specific features to query-conditioned multimodal context. Before fusion, audio carries dense evidence in few tokens and should be preserved, while the more redundant visual stream can be compressed aggressively using local audio--visual structure, as shown in Figure~\ref{fig:teaser}. After fusion, the joint context enables more reliable assessment and pruning of low-relevance tokens across both modalities.

Two observations support our stage-dependent view. First, audio shows higher task-relevant information density and representational diversity per token than video across AVUT and WorldSense (see Figure~\ref{fig:density}). Under a fixed budget, retaining audio consistently performs better, while pruning it yields limited savings and risks losing compact, temporally localized evidence. Second, local audio--visual dynamics capture temporal structure missed by token-wise matching. Neighboring audio and visual representations are informative when they co-evolve within short, aligned windows, whereas pointwise similarity may connect recurring but temporally distant content~\citep{avel,aveclip,localcorr}.

Based on these observations, we propose \textbf{A}udio-\textbf{P}rioritized \textbf{A}llocation with \textbf{C}ontext-aware \textbf{K}V-Cache Pruning (\textbf{\texttt{A-PACK}}), a two-stage token compression framework. Before the LLM, we preserve audio and allocate the visual budget using coarse query relevance and local audio--visual dynamics while removing within-video redundancy. Inside the LLM, we progressively prune low-relevance audio and visual tokens together with their corresponding KV-cache entries. The inner-LLM stage thus realizes deferred audio pruning while reducing computation and KV-cache costs.

Across four omni-modal benchmarks on Qwen2.5-Omni-7B/3B, \ours achieves the strongest average performance among the evaluated training-free methods at both operating points. At the more aggressive setting, it reduces prefill FLOPs by up to $78\%$ while retaining $97.0\%$ of the full-token average performance. It also reduces average GPU memory by $2.3$ GB on AVUT and delivers $2.21\times$ higher decoding throughput during inference. Main contributions are:

\begin{itemize}

    \item We identify an audio--visual asymmetry in per-token utility, motivating audio preservation before the LLM and deferred pruning until query-conditioned interactions.

    \item By using short-window audio--visual dynamics, we guide visual compression beyond token-wise matching while removing within-video redundancy. 

    \item We progressively prune low-relevance audio and visual tokens with their KV-cache entries, reducing prefill, memory, and decoding costs without training.
\end{itemize}

\begin{figure*}[t]
\centering
\includegraphics[width=0.98\textwidth]{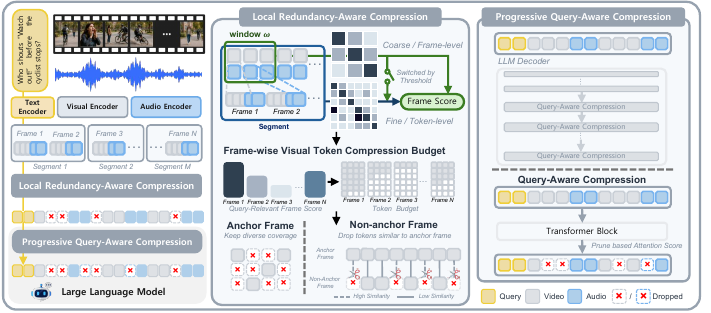}
\caption{Overview of our method. \textbf{Stage~1 (pre-LLM)} preserves audio tokens and compresses only video. It allocates more visual tokens to frames with stronger local audio--visual CKA, then removes within-video redundancy through anchor-based token selection. \textbf{Stage~2 (inner-LLM)} progressively prunes query-irrelevant audio and visual tokens after multimodal interaction. Their corresponding KV-cache entries are also removed, realizing deferred audio pruning and reducing decoding cost.}
\label{fig:method}
\end{figure*}

\section{Related Work}

\subsection{Token Compression for Multimodal LLMs}

The high redundancy and computational cost of long visual sequences have motivated token compression for video LLMs.
Video-LLM token compression is applied either before or inside the language model. Pre-LLM methods reduce visual tokens through temporal segmentation and density pruning~\citep{fastvid} or information-aware allocation~\citep{unicomp}. Inner-LLM methods prune contextualized tokens using attention~\citep{fastv} or text-conditioned relevance~\citep{adaptinfer}. DyCoke further combines early temporal compression with later KV-cache pruning~\citep{dycoke}.

Omni-modal compression additionally requires deciding how audio and video should be treated at different stages. Existing methods use audio saliency~\citep{omnizip}, query-adaptive modality budgets~\citep{omniselect}, asymmetric selection or cross-modal recoverability~\citep{omnisift,contextguard}, or joint audio--visual ranking~\citep{echoingpixels}. These approaches establish the value of modality-aware compression, but most determine modality reduction before fine-grained query-conditioned interactions form. Prior methods primarily optimize which tokens or modalities to reduce before the LLM. In contrast, \ours additionally determines when each modality should be compressed. It preserves audio while reducing visual redundancy before fusion, then switches to joint query-conditioned pruning after fusion.

\subsection{Local Audio--Visual Dynamics}

Audio--visual correspondence has been studied in self-supervised representation learning and sound-source localization, where acoustic signals are associated with their visible sources~\citep{objectsthatsound}. Event-localization methods further identify short intervals in which audible and visible events co-occur~\citep{avel,aveclip}, while locality-aware approaches model temporal continuity to suppress distant or irrelevant matches~\citep{localcorr}.

Representation-level similarity has also been studied with Centered Kernel Alignment (CKA), originally proposed to compare neural representations~\citep{cka}. It has also been adapted for multimodal analysis and distillation~\citep{awcka}. These works primarily use CKA as a training objective or diagnostic measure over global or localized representations. In contrast, \ours applies CKA at inference to short, time-aligned audio--visual windows, capturing local representational co-variation rather than individual token similarity. This signal guides visual-budget allocation together with query relevance, while within-video redundancy is handled separately. This formulation provides a locality-aware, training-free signal for visual token allocation in omni-modal compression.

\section{Method}
We aim to compress long audio--visual inputs by adapting the compression strategy to both modality and processing stage across the LLM boundary, as illustrated in Figure~\ref{fig:method}.

\subsection{Pre-LLM Local Alignment Compression}\label{subsec:prellm}

The pre-LLM stage preserves audio as guidance and focuses compression on video. It estimates local audio--visual correspondence, allocates the visual budget across segments and frames, and retains informative tokens while removing within-scene redundancy.

\paragraph{Alignment as a local, structure-level measure.}
Audio--visual alignment should capture not only token-wise similarity but also whether the modalities \emph{change together} over a short interval. We therefore use Centered Kernel Alignment (CKA)~\citep{cka,awcka} within a temporal window to compare their covariance structures. CKA is symmetric unlike attention, preserves richer structure than mutual-$k$NN, and avoids the Sinkhorn cost and mass constraint of optimal transport, which can force irrelevant matches when visual tokens greatly outnumber audio tokens~\citep{sinkhorn}.

\paragraph{Coarse pass.} We mean-pool the visual tokens of each frame and resample the audio sequence onto the frame axis, producing paired frame-level embeddings $a_f$ and $v_f$. Over a length-$w$ sliding window $\mathcal{W}_f=\{f{-}\lfloor w/2\rfloor,\dots,f{+}\lfloor w/2\rfloor\}$ with $w{=}3$, we stack the mean-centered embeddings into $A_f,V_f\in\mathbb{R}^{w\times d}$. We then compute their linear CKA as the frame-level alignment score: \begin{equation} c_f=\mathrm{CKA}(A_f,V_f)= \frac{\lVert V_f^{\top}A_f\rVert_F^2}{\lVert A_f^{\top}A_f\rVert_F\,\lVert V_f^{\top}V_f\rVert_F}. \label{eq:cka} \end{equation} A larger $c_f$ indicates that the local audio and visual representations exhibit more similar temporal variation. \paragraph{Gated fine pass.} Let $\bar c=\tfrac{1}{F}\sum_f c_f$ denote the mean window CKA. A short-window statistic may miss temporally offset correspondence, such as narration preceding an action or a sound persisting after the visible event. Consequently, $c_f$ can be low even when the two streams remain semantically aligned. When the overall alignment is weak, i.e., $\bar c<\tau$, we supplement $c_f$ with the per-frame audio--visual similarity score $1-\hat d_f\in[0,1]$. Here, $\hat d_f$ is the min--max-normalized direct distance between the visual tokens in frame $f$ and the audio tokens. This term captures correspondence even when the modalities no longer co-vary within the same window: \begin{equation} s_f= \begin{cases} c_f, & \bar c\ge\tau,\\[2pt] (1-\beta)\,c_f+\beta\,(1-\hat d_f), & \bar c<\tau. \end{cases} \label{eq:gate} \end{equation} The weight $\beta$ balances local co-variation and direct correspondence. Thus, the gated score relies on CKA when local alignment is reliable and uses direct similarity as a fallback for temporally shifted events. \paragraph{Segmentation and token selection.} We segment the video using consecutive-frame continuity and allocate the visual budget according to segment importance, following the motivation of CoSeLECT~\citep{coselect}. Let $\mathcal{G}_m$ denote the frames in segment $m$, $r_f$ the query relevance of frame $f$, and $\lambda$ the balance between peak and sustained relevance. We score each segment as \begin{equation} g_m=\sqrt{|\mathcal{G}_m|}\left[\lambda\max_{f\in\mathcal{G}_m}r_f +(1-\lambda)\frac{1}{|\mathcal{G}_m|}\sum_{f\in\mathcal{G}_m}r_f\right]. \label{eq:segment_score} \end{equation} The maximum term preserves brief query-critical moments, while the mean term rewards evidence sustained throughout the segment. The square-root term accounts for segment duration without allowing long segments to dominate the allocation. Given a total visual budget $B_v$, we assign the segment budget as \begin{equation} B_m=B_v\frac{g_m}{\sum_{m'}g_{m'}}, \label{eq:segment_budget} \end{equation} where the denominator normalizes importance across all segments. Within each segment, the frame-level allocation combines query relevance with the local alignment score $s_f$ from Eq.~\eqref{eq:gate}. Both signals reuse cached backbone embeddings, requiring neither recursive search nor an auxiliary encoder~\citep{aks}. After allocating the frame budgets, we remove within-scene redundancy through anchor-based token selection. The first frame of each scene serves as the anchor and retains a diverse, high-coverage token set using density-aware farthest-point sampling (DA-FPS)~\citep{dafps}. DA-FPS favors representative tokens dissimilar to those already selected, yielding compact yet broad visual coverage. For each non-anchor frame, we compare its tokens with anchor tokens at the same spatial positions. We retain the least-similar tokens up to the frame budget, preserving newly appearing evidence while removing content already represented by the anchor.

\begin{table*}[!t]
\centering
\small
\renewcommand{\arraystretch}{0.92}\setlength{\tabcolsep}{3.6pt}
\begin{tabular}{l c cc ccccc}
\toprule
\multirow{2}{*}{Method} & \multirow{2}{*}{$R$ (\%)} &
\multicolumn{2}{c}{Efficiency} & \multicolumn{4}{c}{Benchmark} &
\multirow{2}{*}{Avg. (\%)$\uparrow$} \\
\cmidrule(lr){3-4}\cmidrule(lr){5-8}
& & Prefill FLOPs (T)$\downarrow$ & Final Ret. (\%)$\downarrow$ & AVUT$\uparrow$ & WorldSense$\uparrow$ &
Video-MME$\uparrow$ & DailyOmni$\uparrow$ & \\
\midrule
\multicolumn{9}{c}{\emph{Qwen2.5-Omni-7B}} \\
\midrule
\rowcolor{mycol2!45!white}
Full Tokens & 100 & 67.9 & 100.0 & 64.5 & 46.8 & 66.0 & 63.0 & 100.0 \\
\midrule
Random & 35 & \underline{23.0} & 41.1 & 60.7 & 43.2 & 65.6 & 56.4 & 94.0 \\
DyCoke (V\&A)~\citeyearpar{dycoke} & 35 & 23.3 & 40.9 & 59.9 & 43.0 & 66.0 & 56.1 & 93.6 \\
FlashVID~\citeyearpar{flashvid} & 35 & 23.8 & \underline{27.1} & 61.0 & 44.6 & - & 57.8 & 90.7$\dagger$ \\
FastV~\citeyearpar{fastv} & 35 & 26.0 & 36.5 & 60.2 & 44.1 & - & 59.4 & 90.8$\dagger$ \\
UniComp~\citeyearpar{unicomp} & 35 & 24.2 & 42.0 & 61.5 & 44.3 & 65.9 & 59.9 & 96.3 \\
OmniSelect~\citeyearpar{omniselect} & 35 & 26.3 & 43.5 & \underline{62.9} & \underline{45.6} & 63.6 & \underline{60.0} & 96.5 \\
OmniZip~\citeyearpar{omnizip} & 35 & \textbf{22.9} & 40.2 & 62.4 & 45.3 & \underline{66.6} & 58.9 & \underline{97.0} \\
\rowcolor{mycol1}
\textbf{A-PACK} & 35 & \textbf{22.9} & \textbf{16.8} & \textbf{64.0} & \textbf{46.1} & \textbf{66.7} & \textbf{60.8} & \textbf{98.8} \\
\midrule
Random & 25 & \underline{16.5} & 30.3 & 58.0 & 41.6 & 65.6 & 53.3 & 90.9 \\
DyCoke (V\&A)~\citeyearpar{dycoke} & 25 & 18.5 & 33.2 & 58.3 & 40.3 & 66.1 & 51.6 & 90.0 \\
FlashVID~\citeyearpar{flashvid} & 25 & 17.1 & 23.3 & \underline{60.2} & \underline{43.5} & - & 56.7 & 89.0$\dagger$ \\
FastV~\citeyearpar{fastv} & 25 & 18.5 & \underline{21.2} & 52.1 & 41.0 & 64.0 & 53.4 & 87.6 \\
UniComp~\citeyearpar{unicomp} & 25 & \underline{16.5} & 28.6 & 56.9 & 42.1 & \textbf{66.4} & \underline{58.0} & 92.9 \\
OmniSelect~\citeyearpar{omniselect} & 25 & 21.1 & 34.7 & 58.0 & 40.9 & 65.9 & 53.5 & 90.8 \\
OmniZip~\citeyearpar{omnizip} & 25 & 16.7 & 31.0 & 59.3 & 42.7 & 65.6 & 55.9 & \underline{93.0} \\
\rowcolor{mycol1}
\textbf{A-PACK} & 25 & \textbf{15.0} & \textbf{5.6} & \textbf{62.2} & \textbf{45.3} & \underline{66.2} & \textbf{59.5} & \textbf{97.0} \\
\midrule
\multicolumn{9}{c}{\emph{Qwen2.5-Omni-3B}} \\
\midrule
\rowcolor{mycol2!45!white}
Full Tokens & 100 & 34.9 & 100.0 & 62.5 & 46.2 & 62.6 & 61.6 & 100.0 \\
\midrule
Random & 35 & 11.9 & 44.0 & 58.7 & 43.5 & 61.8 & 54.6 & 93.9 \\
DyCoke (V\&A)~\citeyearpar{dycoke} & 35 & 12.2 & 43.4 & 57.4 & 43.3 & \textbf{62.9} & 53.5 & 93.4 \\
FlashVID~\citeyearpar{flashvid} & 35 & 12.4 & \underline{29.7} & \underline{60.1} & 44.1 & - & 56.4 & 94.3$\dagger$ \\
FastV~\citeyearpar{fastv} & 35 & 11.9 & 36.5 & 56.4 & 43.5 & - & 55.8 & 89.2$\dagger$ \\
UniComp~\citeyearpar{unicomp} & 35 & 11.5 & 41.9 & 59.9 & \underline{44.7} & 62.4 & 57.6 & 96.4 \\
OmniSelect~\citeyearpar{omniselect} & 35 & 12.2 & 42.2 & 58.5 & 44.3 & \underline{62.6} & \textbf{58.9} & 96.3 \\
OmniZip~\citeyearpar{omnizip} & 35 & \underline{10.8} & 40.2 & \underline{60.1} & \textbf{45.3} & 62.4 & 57.2 & \underline{96.7} \\
\rowcolor{mycol1}
\textbf{A-PACK} & 35 & \textbf{10.5} & \textbf{16.8} & \textbf{60.7} & \textbf{45.3} & 62.5 & \underline{57.8} & \textbf{97.3} \\
\midrule
Random & 25 & 8.4 & 33.1 & 56.6 & 41.4 & 61.5 & 52.4 & 91.0 \\
DyCoke (V\&A)~\citeyearpar{dycoke} & 25 & 8.7 & 32.6 & 53.2 & 41.1 & 61.6 & 49.3 & 88.4 \\
FlashVID~\citeyearpar{flashvid} & 25 & 8.8 & 25.2 & \underline{59.2} & \underline{43.8} & - & \underline{56.8} & \underline{93.9}$\dagger$ \\
FastV~\citeyearpar{fastv} & 25 & 8.7 & \underline{24.6} & 51.0 & 41.9 & 60.1 & 52.1 & 88.0 \\
UniComp~\citeyearpar{unicomp} & 25 & \underline{7.7} & 28.6 & 54.6 & 43.1 & 60.4 & 54.0 & 91.1 \\
OmniSelect~\citeyearpar{omniselect} & 25 & 8.7 & 31.3 & 55.1 & 42.5 & 61.6 & 55.3 & 92.1 \\
OmniZip~\citeyearpar{omnizip} & 25 & \underline{7.7} & 31.0 & 57.0 & 42.1 & \underline{61.8} & 54.2 & 92.4 \\
\rowcolor{mycol1}
\textbf{A-PACK} & 25 & \textbf{6.7} & \textbf{5.6} & \textbf{59.3} & \textbf{44.0} & \textbf{61.9} & \textbf{57.3} & \textbf{95.5} \\
\bottomrule
\end{tabular}
\caption{Main results on four multimodal benchmarks for Qwen2.5-Omni-7B and 3B.
The best and second-best compressors are bold and underlined. DyCoke (V\&A) applies only its TTM module separately to audio and video; Random, OmniZip, and
OmniSelect also compress both modalities. `-' denotes an out-of-memory (OOM) result. $\dagger$ denotes a three-benchmark mean.}
\label{tab:main7b}
\end{table*}

\begin{table}[]
\centering
\setlength{\tabcolsep}{2pt}
\renewcommand{\arraystretch}{1.0}
{\small
\begin{tabularx}{\columnwidth}{@{}l*{4}{>{\raggedleft\arraybackslash}X}@{}}
\toprule
\multirow{2}{*}{Method} & \multicolumn{2}{c}{Resource and Quality} &
\multicolumn{2}{c}{Inference Speed} \\
\cmidrule(lr){2-3}\cmidrule(lr){4-5}
& \shortstack{Memory\\(GB)$\downarrow$} &
\shortstack{Accuracy\\(\%)$\uparrow$} &
\shortstack{Prefill\\Speedup$\uparrow$} &
\shortstack{End-to-end\\Speedup$\uparrow$} \\
\midrule
Full & 24.0 & 64.5 & 1.00$\times$ & 1.00$\times$ \\
FlashVID & 22.5 & \underline{60.2} & 1.65$\times$ & 1.73$\times$ \\
DyCoke & 22.5 & 58.3 & 1.67$\times$ & 1.77$\times$ \\
FastV & 22.6 & 52.1 & \underline{1.80}$\times$ & \underline{1.89}$\times$ \\
UniComp & 22.4 & 56.9 & 1.68$\times$ & 1.77$\times$ \\
OmniZip & \underline{21.9} & 59.3 & 1.75$\times$ & 1.86$\times$ \\
\rowcolor{mycol1}
\textbf{A-PACK} & \textbf{21.7} & \textbf{62.2} & \textbf{1.85}$\times$ & \textbf{1.96}$\times$ \\
\bottomrule
\end{tabularx}%
}
\caption{Efficiency on AVUT with Qwen2.5-Omni-7B at the $25\%$ prefill-FLOPs tier. A-PACK achieves the highest accuracy, lowest memory use, and fastest prefill and end-to-end inference.}
\label{tab:eff-avut}
\end{table}

\subsection{Inner-LLM Query-Aware Compression}\label{subsec:innerllm}

Pre-LLM representations do not yet contain fine-grained query-conditioned multimodal context. We therefore defer audio pruning until the decoder has integrated audio, video, and the query. Starting from a middle decoder layer, A-PACK progressively reassesses the relevance of retained audio and visual tokens using query attention. At a pruning layer, let $\mathcal{S}$ denote the currently cached multimodal tokens, $q$ the representation of the last query token, $k_i$ the key of token $i$, and $P$ the pruning ratio. We score each token using scaled query--key attention: \begin{equation} \displaystyle I_i= \frac{\exp\!\left(q^\top k_i/\sqrt{d}\right)} {\sum_{j\in\mathcal{S}}\exp\!\left(q^\top k_j/\sqrt{d}\right)}, \qquad i\in\mathcal{S}. \label{eq:inner_score} \end{equation} Here, $d$ is the key dimension. For a multi-head decoder, we average the score across attention heads. A larger $I_i$ indicates greater relevance to the query, and the score requires no additional forward pass because the query--key interactions are already computed by the decoder. We then retain the highest-scoring $(1-P)$ fraction of the current tokens: \begin{equation} \displaystyle \mathcal{S}\leftarrow \operatorname{TopK}_{\lceil(1-P)|\mathcal{S}|\rceil} \left(\mathcal{S};I\right). \label{eq:inner} \end{equation} Pruning begins after early decoder layers establish multimodal context and is repeated at subsequent pruning layers as the representations become more query-conditioned. Starting from $N_0$ cached tokens, each step retains a $(1-P)$ fraction of the current sequence. The remaining tokens and their corresponding key--value entries are removed, reducing attention computation in later layers and throughout autoregressive decoding.

\section{Experiments}
\subsection{Experimental Setup}

\paragraph{Benchmarks.}
We evaluate on five audio--visual understanding benchmarks: \textbf{AVUT}~\citep{avut}, \textbf{WorldSense}~\citep{worldsense}, \textbf{DailyOmni}~\citep{dailyomni}, \textbf{Video-MME}~\citep{videomme}, and \textbf{AVHBench}~\citep{avhbench}. AVUT is audio-centric, while WorldSense covers joint audio--visual understanding across eight domains. DailyOmni focuses on everyday audio--visual events, and Video-MME evaluates general video understanding without subtitles. AVHBench provides balanced binary judgments for audio--visual hallucination.

\paragraph{Comparison Methods.}
We compare A-PACK with representative training-free token compressors: FastV~\citep{fastv}, DyCoke~\citep{dycoke}, OmniZip~\citep{omnizip}, UniComp~\citep{unicomp}, FlashVID~\citep{flashvid}, OmniSelect~\citep{omniselect}, and Random.
All baselines use the same backbone without fine-tuning. Full Tokens denotes the uncompressed model. 
DyCoke applies its original temporal token merging separately to audio and video. FastV computes attention at layer~$3$ before pruning visual tokens. 
We use the released selection/compression cores of UniComp, OmniSelect, and FlashVID, with method-specific settings adjusted for the target $R$ tiers. OmniZip uses its released setting at the $35\%$ tier, for the $25\%$ tier, we set both ratios to $0.75$ for more aggressive compression.
\paragraph{Implementation Details.} For fair comparison, we implement all models on the same Qwen2.5-Omni~\citep{qwen25omni} 3B/7B backbone and use NVIDIA A6000 48 \,GB GPUs without fine-tuning. 
For each backbone, Table~\ref{tab:main7b} reports matched $35\%$ and $25\%$ prefill-FLOPs tiers. $R$ is the target prefill-FLOPs ratio relative to Full Tokens and Final Ret. is the ratio of tokens remaining at the end of prefill after inner-LLM pruning. A-PACK starts inner-LLM pruning at layers $14$/$18$ for 7B/3B using a $10\%$/$15\%$ per-layer drop in the $35\%$/$25\%$ tiers. Hyperparameter settings are listed in Supplementary Table~\ref{tab:hparams}; ablations fix the inner-LLM per-layer drop at $P{=}15\%$, while varying only the parameter under study. 
We report accuracy and efficiency under the common protocol in Supplementary Section~G.

\begin{figure*}[]
\centering
{
\begin{tabular}{@{}cc@{}}
\includegraphics[width=0.570\textwidth]{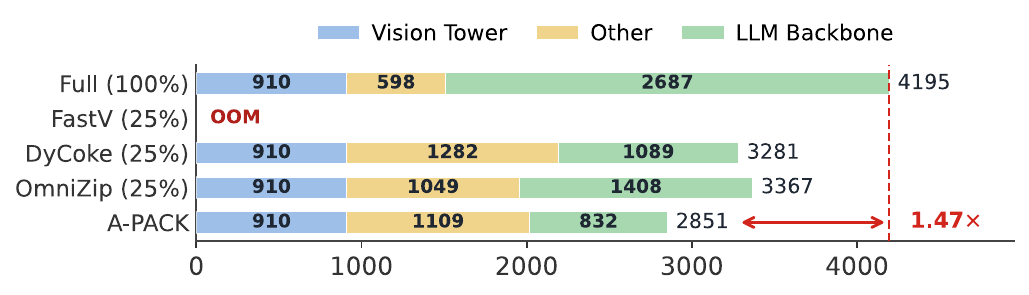} &
\includegraphics[width=0.405\textwidth]{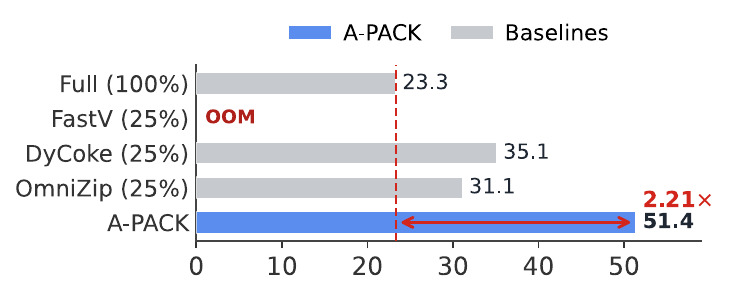} \\
\textbf{(a)} Time-to-first-token (TTFT) breakdown & \textbf{(b)} Decoding throughput
\end{tabular}}
\caption{Inference efficiency of Qwen2.5-Omni-7B at the $25\%$ prefill-FLOPs tier. (a) TTFT breakdown into the vision tower, preprocessing with pre-LLM compression, and the LLM backbone with inner-LLM compression. (b) Decoding throughput.}
\label{fig:infspeed}
\end{figure*}

\begin{figure}[]
\centering
\includegraphics[width=0.92\columnwidth]{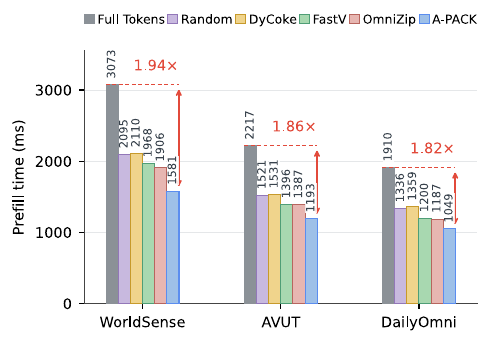}
\caption{A-PACK has the lowest prefilling time on omni-modal benchmarks owing to its inner-LLM stage.}
\label{fig:prefill}
\end{figure}

\begin{figure*}[]
\centering
\begin{subfigure}[t]{0.30\textwidth}
\centering
\includegraphics[width=\linewidth]{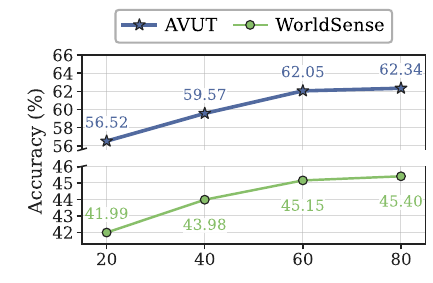}
\caption{Audio kept ratio (\%)}
\label{fig:audiopruning}
\end{subfigure}\hfill
\begin{subfigure}[t]{0.30\textwidth}
\centering
\includegraphics[width=\linewidth]{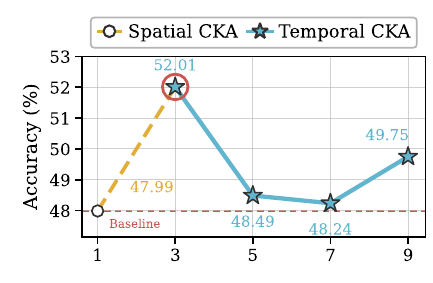}
\caption{Window-size sensitivity}
\label{fig:hparamsens}
\end{subfigure}\hfill
\begin{subfigure}[t]{0.30\textwidth}
\centering
\includegraphics[width=\linewidth]{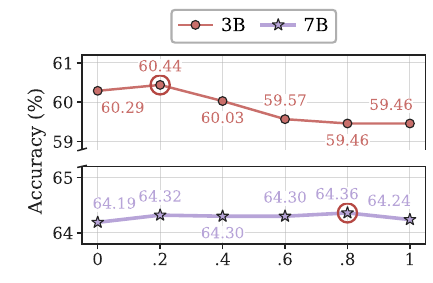}
\caption{CKA alignment weight}
\label{fig:ckaweight}
\end{subfigure}
\caption{Key sensitivity analyses. 
(a) Retaining more audio improves accuracy under a fixed total budget.
(b) WorldSense accuracy peaks with a short temporal window.
(c) Performance remains stable across the tested alignment weights. Red circle indicates highest score. Baseline indicates per-token matching score accuracy.}
\label{fig:key-sweeps}
\end{figure*}

\subsection{Main Results}
We use LMMs-Eval~\citep{lmmseval} for Video-MME and a unified evaluation pipeline for the other benchmarks. Avg. denotes the mean relative accuracy across benchmarks, with Full Tokens normalized to $100\%$. Table~\ref{tab:main7b} compares methods at the $35\%$ and $25\%$ prefill-FLOPs tiers on Qwen2.5-Omni-7B and 3B.
A-PACK achieves the highest average accuracy across both model scales and operating points. On the 7B model, it retains $98.8\%$ of Full-Tokens accuracy at the $35\%$ tier while reducing prefill FLOPs by about $66\%$. At the more aggressive $25\%$ tier, it retains $97.0\%$ while reducing prefill FLOPs by about $78\%$ and final token retention to $5.6\%$. The gains over OmniZip are largest on AVUT and DailyOmni, supporting the use of preserved audio to guide visual compression. The 3B results show the same trend, retaining $97.3\%$ and $95.5\%$ average accuracy at the two tiers, respectively.

\subsection{Efficiency Analyses}
Table~\ref{tab:eff-avut} and Supplementary Table~\ref{tab:eff-ws} report the
Qwen2.5-Omni-7B efficiency results at the $25\%$ prefill-FLOPs tier.
Because the inner-LLM stage shortens the KV cache, it also accelerates decoding, unlike methods that prune tokens only before the LLM.

\begin{figure*}[]
\centering
{\small
\begin{tabular}{@{}cccc@{}}
\begin{tabular}[t]{@{}c@{}}
\includegraphics[width=0.172\textwidth]{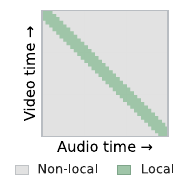} \\
\textbf{(a)} Local-window mask
\end{tabular} &
\begin{tabular}[t]{@{}c@{}}
\includegraphics[width=0.153\textwidth]{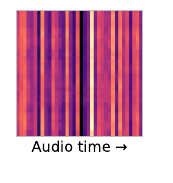} \\
\textbf{(b)} Static cosine
\end{tabular} &
\begin{tabular}[t]{@{}c@{}c@{}}
\includegraphics[width=0.123\textwidth]{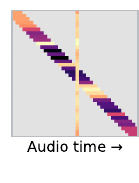} &
\includegraphics[width=0.079\textwidth]{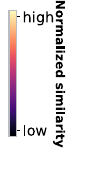} \\
\parbox[t]{0.123\textwidth}{\centering\textbf{(c)} A-PACK} &
\parbox[t]{0.079\textwidth}{\centering\strut}
\end{tabular} &
\begin{tabular}[t]{@{}c@{}}
\includegraphics[width=0.399\textwidth]{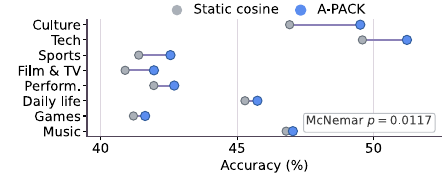} \\
\textbf{(d)} Domain accuracy: static vs. A-PACK
\end{tabular}
\end{tabular}}
\caption{\textbf{(a)} A local-window mask defines locally aligned audio--visual token pairs.
\textbf{(b)} Static cosine can assign high similarity to temporally distant matches.
\textbf{(c)} A-PACK uses gated local temporal correspondence to guide visual selection.
\textbf{(d)} Local-window alignment improves all eight WorldSense domains; the improvement is
statistically significant (exact McNemar $p{=}0.0117$).}
\label{fig:confusion}
\end{figure*}
Figure~\ref{fig:infspeed} breaks down TTFT and reports decoding throughput. A-PACK has slightly higher pre-LLM overhead than OmniZip due to local audio--visual alignment. However, progressive inner-LLM pruning shortens backbone time and the KV cache, reaching $51.4$ tok/s versus $31.1$ for OmniZip and $35.1$ for DyCoke. It also achieves the lowest prefill time across benchmarks (Figure~\ref{fig:prefill}). DyCoke has greater pre-LLM overhead and a longer cache, while FastV runs out of memory by attending to the full visual sequence before layer~3. The 3B model shows the same trend (Supplementary Figure~\ref{fig:prefill3b}).

\subsection{Ablation Study on Compression Stages}
Across 3B and 7B, A-PACK pre-LLM selection with a random inner stage outperforms random pre-LLM selection with A-PACK inner pruning on both AVUT and DailyOmni (Table~\ref{tab:stagecombo-7b}; Supplementary Table~\ref{tab:stagecombo-3b}). Thus, the pre-LLM stage provides most of the accuracy gain, while query-aware inner pruning recovers the remaining performance and reduces the later sequence cost. Using A-PACK in both stages is best.

\begin{table}[t]
\centering
\renewcommand{\arraystretch}{0.95}
\setlength{\tabcolsep}{2pt}
{\small
\begin{tabularx}{\columnwidth}{@{}cc*{3}{>{\centering\arraybackslash}X}@{}}
\toprule
\multicolumn{2}{c}{Method} & AVUT & DailyOmni &
\multirow{2}{*}{FLOPs$\downarrow$} \\
\cmidrule(lr){1-2}\cmidrule(lr){3-4}
Pre-LLM & Inner-LLM & Acc. (\%) & Acc. (\%) & \\
\midrule
Full & Full & 64.5 & 63.0 & 100\% \\
Random & Random & 61.5 & 56.6 & 38.3\% \\
Random & A-PACK & 61.6 & 56.4 & \textbf{32.4}\% \\
A-PACK & Random & \underline{63.3} & \underline{58.4} & 38.2\% \\
A-PACK & FastV & 62.6 & 57.9 & \underline{32.7}\% \\
A-PACK & DyCoke & 63.0 & 57.8 & 39.8\% \\
\rowcolor{mycol1}
\textbf{A-PACK} & \textbf{A-PACK} & \textbf{64.2} & \textbf{58.5} & \textbf{32.4}\% \\
\bottomrule
\end{tabularx}%
}
\caption{Stage-combination ablation on Qwen2.5-Omni-7B. All non-base rows use a $50\%$
pre-LLM retention ratio and the same inner budget (per-layer $P{=}15\%$).
}
\label{tab:stagecombo-7b}
\end{table}

\subsection{Ablation Study on Pre-LLM Compression}

\paragraph{Audio carries denser information per token.}
Figure~\ref{fig:density} shows that audio carries more task-relevant
  information per token than video: masking an audio token changes the answer
  distribution $9.8\times$ more, and audio representations have $1.7\times$
  higher effective rank. We therefore preserve audio as guidance and meet the
  pre-LLM budget by compressing video; Supplementary Section~D gives the measurement protocol.

\paragraph{Pre-LLM retention ratio.}
A retention sweep (Supplementary Figure~\ref{fig:preratio}) traces the
pre-LLM accuracy--cost trade-off from $30\%$ to $50\%$. 
Accuracy rises with retention, so for each tier we choose the highest pre-LLM ratio that remains within its prefill-FLOPs budget: $50\%$ for the $35\%$ tier and $35\%$ for the $25\%$ tier.

\begin{table}[t]
\centering
\renewcommand{\arraystretch}{0.95}
\setlength{\tabcolsep}{1.0pt}
{\small
\begin{tabular}{@{}ll*{6}{c}@{}}
\toprule
\multirow{2}{*}{Type} &
\multirow{2}{*}{Method} &
\multicolumn{5}{c}{AVHBench} &
\multicolumn{1}{c}{WorldSense} \\
\cmidrule(lr){3-7}\cmidrule(lr){8-8}
& & Acc. $\uparrow$ & Prec. $\uparrow$ & Rec. $\uparrow$ &
F1 $\uparrow$ & Yes $\uparrow$ & Acc. $\uparrow$ \\
\midrule

\multirow{4}{*}{Base}
& Base
& 73.6 & 72.7 & \underline{75.5} & \underline{74.1}
& \underline{51.9} & 44.9 \\
& Random
& 73.6 & \underline{73.0} & 74.8 & 73.9
& 51.3 & 44.3 \\
& Attention
& 73.4 & 72.9 & 74.5 & 73.7
& 51.0 & 44.6 \\
& Audio Sal.
& 73.5 & 72.9 & 74.7 & 73.8
& 51.2 & 44.5 \\

\midrule
Sim.
& AV Cosine
& \underline{73.7} & \textbf{73.3} & 74.5 & 73.9
& 50.9 & \underline{44.7} \\

\midrule
\multirow{2}{*}{Align.}
& OT
& 73.4 & 72.9 & 74.5 & 73.7
& 51.1 & 44.3 \\
& CKA\,$+$\,OT
& 73.2 & 72.7 & 74.3 & 73.5
& 51.1 & 44.6 \\

\midrule
\rowcolor{mycol1}
\textbf{Ours}
& \textbf{Gated CKA}
& \textbf{74.0} & 72.7 & \textbf{77.0} & \textbf{74.8}
& \textbf{53.0} & \textbf{46.2} \\
\bottomrule
\end{tabular}%
}

\parbox{\linewidth}{%
\footnotesize\raggedright 
Sim.: similarity \hspace{0.8em}
Align.: alignment \\
Audio Sal.: audio saliency \hspace{0.8em}
OT: optimal transport
}

\caption{Alignment-scoring comparison on AVHBench and WorldSense under a fixed compression setting. Acc. denotes accuracy (\%); Prec. denotes precision (\%); Rec. denotes recall (\%); F1 denotes F1 score (\%); and Yes denotes the proportion of ``Yes'' predictions (\%).}
\label{tab:alignment}
\end{table}

\paragraph{Retained audio helps audio- and visual-centric tasks.}
  Fixing the total-token budget, we vary the \emph{audio kept ratio} and
  measure accuracy on AVUT and WorldSense. Figure~\ref{fig:audiopruning} shows
  monotonic gains on both benchmarks: raising audio retention from
  $20\%$ to $80\%$ improves relative accuracy by about $10\%$ on AVUT and
  $8\%$ on WorldSense. Thus, even on the comparatively visual WorldSense,
  audio acts as useful guidance rather than competing for the token budget.
  An LLM judge characterizes AVUT as audio-centric and WorldSense as
  visual-centric (Supplementary Figures~\ref{fig:querymod} and \ref{fig:audiocentric}).

\subsection{Ablation Study on Hyperparameters}
\paragraph{Hyperparameter sensitivity.}
Figure~\ref{fig:hparamsens} compares spatial CKA at $w{=}1$, which measures
within-frame visual-patch covariance and the co-timed audio interval, with temporal
CKA over larger windows. The result is highest at $w{=}3$, suggesting that
a local history of cross-modal change is useful. Gate-threshold and fine-blend sweeps
are reported in Supplementary Figures~\ref{fig:hparam-gate} and
\ref{fig:hparam-blend}; the gate is more sensitive, whereas the fine-blend weight
is comparatively flat around $\beta{=}0.90$.

\paragraph{Alignment-score weight.}
Figure~\ref{fig:ckaweight} sweeps the CKA weight from $0$ (pure visual importance) to $1$ (pure CKA) on AVUT. In the 7B model, the score varies by less than $0.2$ points over the sweep, whereas 3B peaks at a small CKA weight and declines toward pure CKA. Alignment therefore acts as a complementary correction to visual importance rather than a signal that should dominate it.
\subsection{Analysis of Audio-Visual Alignment}
\paragraph{Alignment-scoring alternatives.}\label{subsec:local}
On AVHBench's balanced binary judgments, our Gated CKA improves recall and F1 over AV cosine by $2.5$ and $0.9$ points, respectively (Table~\ref{tab:alignment}), indicating greater robustness to audio--visual hallucinations. It also achieves the best WorldSense accuracy (see Supplementary Section~E).

\paragraph{Local-correspondence validation.}\label{subsec:alignment-eval}
Figure~\ref{fig:confusion} shows why local correspondence is needed. Static cosine can make temporally separated events appear aligned, whereas our method 
retains locally aligned audio--visual dynamics using gated local CKA.  On WorldSense, it achieves higher accuracy in all eight domains and significantly more paired corrections than degradations (exact McNemar $p{=}0.0117$), supporting temporal locality as a useful selection signal.

\section{Conclusion}
We frame omni-modal token compression as a stage- and modality-dependent problem. \ours preserves information-dense audio while reducing redundant visual content before the LLM, then progressively prunes low-relevance multimodal tokens and their KV-cache entries after query-conditioned interactions emerge. Local audio--visual dynamics further provide a stronger cue for visual selection than token-wise similarity. Across four primary benchmarks, \ours reduces prefill FLOPs by up to $78\%$, improves decoding throughput by up to $2.21\times$, and lowers AVUT GPU memory by $2.3$\,GB. These results highlight stage-aware compression across the LLM boundary as a promising principle for efficient omni-modal LLMs.
As a training-free approach, our method improves efficiency, though aggressive compression may affect fine-grained or long-range evidence. Its audio-preserving strategy may vary across models and tasks, while local co-variation can be sensitive to temporal offsets. Future work will explore adaptive retention, training-based extensions, and broader omni-modal architectures.
\bibliography{aaai2027}

@String{CVPR    = "Proceedings of the IEEE/CVF Conference on Computer Vision and Pattern Recognition (CVPR)"}

@String{ECCV    = "Proceedings of the European Conference on Computer Vision (ECCV)"}

@String{ICML    = "Proceedings of the International Conference on Machine Learning (ICML)"}

@String{ICLR    = "Proceedings of the International Conference on Learning Representations (ICLR)"}

@String{NEURIPS = "Advances in Neural Information Processing Systems (NeurIPS)"}

@String{EMNLP   = "Proceedings of the Conference on Empirical Methods in Natural Language Processing (EMNLP)"}

@String{ICASSP  = "Proceedings of the IEEE International Conference on Acoustics, Speech, and Signal Processing (ICASSP)"}

@String{TMLR    = "Transactions on Machine Learning Research (TMLR)"}

@String{TCSVT    = "IEEE Transactions on Circuits and Systems for Video Technology (TCSVT)"}

@String{WACV = "Proceedings of the IEEE/CVF Winter Conference on Applications of Computer Vision (WACV)"}

@misc{qwen25omni,
  author = "Xu, Jin and Guo, Zhifang and He, Jinzheng and Hu, Hangrui and He, Ting and Bai, Shuai and Chen, Keqin and Wang, Jialin and Fan, Yang and Dang, Kai and others",
  title = "Qwen2.5-Omni Technical Report",
  year = 2025,
  eprint = "2503.20215",
  archivePrefix = "arXiv",
}

@article{localcorr,
  title={Locality-Aware Cross-Modal Correspondence Learning for Dense Audio-Visual Events Detection},
  author={Xing, Ling and Qu, Hongyu and Yan, Rui and Shu, Xiangbo and Tang, Jinhui},
  journal=TCSVT,
  volume={36},
  number={5},
  pages={6838--6851},
  year={2026},
  doi={10.1109/TCSVT.2025.3629609}
}

@inproceedings{aveclip,
  title={AVE-CLIP: AudioCLIP-Based Multi-Window Temporal Transformer for Audio Visual Event Localization},
  author={Mahmud, Tanvir and Marculescu, Diana},
  booktitle=WACV,
  pages={5158--5167},
  year={2023}
}

@article{dafps,
  title={Density-Aware Farthest Point Sampling},
  author={Paolo Climaco and Jochen Garcke},
  journal={Trans. Mach. Learn. Res.},
  year={2025},
  volume={2025},
  url={https://api.semanticscholar.org/CorpusID:281325534}
}

@inproceedings{avel,
  title={Audio-Visual Event Localization in Unconstrained Videos},
  author={Tian, Yapeng and Shi, Jing and Li, Bochen and Duan, Zhiyao and Xu, Chenliang},
  booktitle=ECCV,
  year={2018}
}

@inproceedings{lmmseval,
    title = "{LMM}s-Eval: Reality Check on the Evaluation of Large Multimodal Models",
    author = "Zhang, Kaichen  and
      Li, Bo  and
      Zhang, Peiyuan  and
      Pu, Fanyi  and
      Cahyono, Joshua Adrian  and
      Hu, Kairui  and
      Liu, Shuai  and
      Zhang, Yuanhan  and
      Yang, Jingkang  and
      Li, Chunyuan  and
      Liu, Ziwei",
    editor = "Chiruzzo, Luis  and
      Ritter, Alan  and
      Wang, Lu",
    booktitle = "Findings of the Association for Computational Linguistics: NAACL 2025",
    month = apr,
    year = "2025",
    address = "Albuquerque, New Mexico",
    publisher = "Association for Computational Linguistics",
    url = "https://aclanthology.org/2025.findings-naacl.51/",
    doi = "10.18653/v1/2025.findings-naacl.51",
    pages = "881--916",
    ISBN = "979-8-89176-195-7"
}

@misc{ola,
  author = "Liu, Zuyan and Dong, Yuhao and Wang, Jiahui and Liu, Ziwei and Hu, Winston and Lu, Jiwen and Rao, Yongming",
  title = "Ola: Pushing the Frontiers of Omni-Modal Language Model",
  year = 2025,
  eprint = "2502.04328",
  archivePrefix = "arXiv",
}

@misc{baichuanomni,
  author = "Li, Yadong and Liu, Jun and Zhang, Tao and Chen, Song and Li, Tianpeng and Li, Zehuan and Liu, Lijun and Ming, Lingfeng and Dong, Guosheng and Pan, Da and others",
  title = "Baichuan-Omni-1.5 Technical Report",
  year = 2025,
  eprint = "2501.15368",
  archivePrefix = "arXiv",
}

@inproceedings{llavaonevision,
  author = "Li, Bo and Zhang, Yuanhan and Guo, Dong and Zhang, Renrui and Li, Feng and Zhang, Hao and Zhang, Kaichen and Zhang, Peiyuan and Li, Yanwei and Liu, Ziwei and Li, Chunyuan",
  title = "{LLaVA-OneVision}: Easy Visual Task Transfer",
  booktitle = TMLR,
  year = 2025,
}

@misc{videollama2,
  author = "Cheng, Zesen and Leng, Sicong and Zhang, Hang and Xin, Yifei and Li, Xin and Chen, Guanzheng and Zhu, Yongxin and Zhang, Wenqi and Luo, Ziyang and Zhao, Deli and Bing, Lidong",
  title = "{VideoLLaMA} 2: Advancing Spatial-Temporal Modeling and Audio Understanding in Video-LLMs",
  year = 2024,
  eprint = "2406.07476",
  archivePrefix = "arXiv",
}

@misc{videochatflash,
  author = "Li, Xinhao and Wang, Yi and Yu, Jiashuo and Zeng, Xiangyu and Zhu, Yuhan and Huang, Haian and Gao, Jianfei and Li, Kunchang and He, Yinan and Wang, Chenting and others",
  title = "{VideoChat-Flash}: Hierarchical Compression for Long-Context Video Modeling",
  year = 2025,
  eprint = "2501.00574",
  archivePrefix = "arXiv",
}

@inproceedings{longvila,
  author = "Chen, Yukang and Xue, Fuzhao and Li, Dacheng and Hu, Qinghao and Zhu, Ligeng and Li, Xiuyu and Fang, Yunhao and Tang, Haotian and Yang, Shang and Liu, Zhijian and others",
  title = "{LongVILA}: Scaling Long-Context Visual Language Models for Long Videos",
  booktitle = ICLR,
  year = 2025,
}

@inproceedings{longvu,
  author = "Shen, Xiaoqian and Xiong, Yunyang and Zhao, Changsheng and Wu, Lemeng and Chen, Jun and Zhu, Chenchen and Liu, Zechun and Xiao, Fanyi and Varadarajan, Balakrishnan and Bordes, Florian and others",
  title = "{LongVU}: Spatiotemporal Adaptive Compression for Long Video-Language Understanding",
  booktitle = ICML,
  year = 2025,
}

@inproceedings{fastv,
  author = "Chen, Liang and Zhao, Haozhe and Liu, Tianyu and Bai, Shuai and Lin, Junyang and Zhou, Chang and Chang, Baobao",
  title = "An Image Is Worth 1/2 Tokens After Layer 2: Plug-and-Play Inference Acceleration for Large Vision-Language Models",
  booktitle = ECCV,
  year = 2024,
}

@inproceedings{dycoke,
  author = "Tao, Keda and Qin, Can and You, Haoxuan and Sui, Yang and Wang, Huan",
  title = "{DyCoke}: Dynamic Compression of Tokens for Fast Video Large Language Models",
  booktitle = CVPR,
  year = 2025,
}

@inproceedings{fastvid,
  author = "Shen, Leqi and Gong, Guoqiang and He, Tao and Zhang, Yifeng and Liu, Pengzhang and Zhao, Sicheng and Ding, Guiguang",
  title = "{FastVID}: Dynamic Density Pruning for Fast Video Large Language Models",
  booktitle = NEURIPS,
  year = 2025,
}

@inproceedings{holitom,
  author = "Shao, Kele and Tao, Keda and Qin, Can and You, Haoxuan and Sui, Yang and Wang, Huan",
  title = "{HoliTom}: Holistic Token Merging for Fast Video Large Language Models",
  booktitle = NEURIPS,
  year = 2025,
}

@inproceedings{omnizip,
  author = "Tao, Keda and Shao, Kele and Yu, Bohan and Wang, Weiqiang and Liu, Jian and Wang, Huan",
  title = "{OmniZip}: Audio-Guided Dynamic Token Compression for Fast Omnimodal Large Language Models",
  booktitle = CVPR,
  year = 2026,
}

@inproceedings{avhbench,
  author = "Sung-Bin, Kim and Hyun-Bin, Oh and Lee, JungMok and Senocak, Arda and Chung, Joon Son and Oh, Tae-Hyun",
  title = "{AVHBench}: A Cross-Modal Hallucination Benchmark for Audio-Visual Large Language Models",
  booktitle = ICLR,
  year = 2025,
}

@misc{contextguard,
  author = "Jung, Chaeyoung and Rho, Kyeongha and Chung, Joon Son",
  title = "Keep What Audio Cannot Say: Context-Preserving Token Pruning for Omni-LLMs",
  year = 2026,
  eprint = "2605.11605",
  archivePrefix = "arXiv",
}

@inproceedings{cka,
  author = "Kornblith, Simon and Norouzi, Mohammad and Lee, Honglak and Hinton, Geoffrey",
  title = "Similarity of Neural Network Representations Revisited",
  booktitle = ICML,
  year = 2019,
}

@article{erank,
  author = "Roy, Olivier and Vetterli, Martin",
  title = "The Effective Rank: A Measure of Effective Dimensionality",
  journal = "European Signal Processing Conference (EUSIPCO)",
  pages = "606--610",
  year = 2007,
}

@inproceedings{sinkhorn,
  author = "Cuturi, Marco",
  title = "Sinkhorn Distances: Lightspeed Computation of Optimal Transport",
  booktitle = NEURIPS,
  year = 2013,
}

@inproceedings{awcka,
  author = "Yang, Qingran and Zhao, Botao and Kang, Zuheng and Li, Xue and He, Yayun and Liu, Chuhang and Zhang, Xulong and Qu, Xiaoyang and Peng, Junqing and Wang, Jianzong",
  title = "Attention-Weighted Centered Kernel Alignment for Knowledge Distillation in Large Audio-Language Models Applied to Speech Emotion Recognition",
  booktitle = ICASSP,
  year = 2026,
}

@inproceedings{aks,
  author = "Tang, Xi and Qiu, Jihao and Xie, Lingxi and Tian, Yunjie and Jiao, Jianbin and Ye, Qixiang",
  title = "Adaptive Keyframe Sampling for Long Video Understanding",
  booktitle = CVPR,
  year = 2025,
}

@inproceedings{coselect,
  author = "Devnani, Bhavika Suresh and Jain, Jitesh and Shi, Humphrey and Hoffman, Judy",
  title = "{CoSeLECT}: Adaptive Frame Selection for Video-Language Understanding",
  booktitle = "Second Workshop on Video Large Language Models ({VidLLMs}), {CVPR} 2026",
  month = jun,
  year = 2026,
  url = "https://openreview.net/forum?id=KeyLHsSS1N",
  note = "Poster",
}

@inproceedings{videomme,
  author = "Fu, Chaoyou and Dai, Yuhan and Luo, Yongdong and Li, Lei and Ren, Shuhuai and Zhang, Renrui and Wang, Zihan and Zhou, Chenyu and Shen, Yunhang and Zhang, Mengdan and others",
  title = "Video-MME: The First-Ever Comprehensive Evaluation Benchmark of Multi-Modal LLMs in Video Analysis",
  booktitle = CVPR,
  year = 2025,
}

@inproceedings{worldsense,
  author = "Hong, Jack and Yan, Shilin and Cai, Jiayin and Jiang, Xiaolong and Hu, Yao and Xie, Weidi",
  title = "{WorldSense}: Evaluating Real-World Omnimodal Understanding for Multimodal LLMs",
  booktitle = ICLR,
  year = 2026,
}

@misc{dailyomni,
  author = "Zhou, Ziwei and Wang, Rui and Wu, Zuxuan and Jiang, Yu-Gang",
  title = "{Daily-Omni}: Towards Audio-Visual Reasoning with Temporal Alignment Across Modalities",
  year = 2025,
  eprint = "2505.17862",
  archivePrefix = "arXiv",
}

@inproceedings{avut,
  author = "Yang, Yudong and Zhuang, Jimin and Sun, Guangzhi and Tang, Changli and Li, Yixuan and Li, Peihan and Jiang, Yifan and Li, Wei and Ma, Zejun and Zhang, Chao",
  title = "Audio-Centric Video Understanding Benchmark without Text Shortcut",
  booktitle = EMNLP,
  year = 2025,
}

@InProceedings{unicomp,
    author    = {Yuan, Chao and Chen, Shimin and Lin, Minliang and Qiao, Limeng and Wan, Guanglu and Ma, Lin},
    title     = {UniComp: Rethinking Video Compression Through Informational Uniqueness},
    booktitle = {Proceedings of the IEEE/CVF Conference on Computer Vision and Pattern Recognition (CVPR)},
    month     = {June},
    year      = {2026},
    pages     = {18609-18618}
}

@inproceedings{flashvid,
  author = "Fan, Ziyang and Chen, Keyu and Xing, Ruilong and Li, Yulin and Jiang, Li and Tian, Zhuotao",
  title = "{FlashVID}: Efficient Video Large Language Models via Training-Free Tree-Based Spatiotemporal Token Merging",
  booktitle = ICLR,
  year = 2026,
}

@misc{adaptinfer,
  author = "Zhang, Weichen and Zhu, Zhui and Li, Ningbo and Tao, Shilong and Liu, Kebin and Liu, Yunhao",
  title = "{AdaptInfer}: Adaptive Token Pruning for Vision-Language Model Inference with Dynamical Text Guidance",
  year = 2025,
  eprint = "2508.06084",
  archivePrefix = "arXiv",
}

@misc{omniselect,
  author = "Yang, Morunliu and Xu, Ruotao and Li, Le and Wang, Yue and Zhang, Jianxin and Li, Juntao and Lou, Yihang and Feng, Siwei and Li, Peifeng",
  title = "{OmniSelect}: Dynamic Modality-Aware Token Compression for Efficient Omni-Modal Large Language Models",
  year = 2026,
  eprint = "2605.18041",
  archivePrefix = "arXiv",
}

@inproceedings{echoingpixels,
  author = "Gong, Chao and Wang, Depeng and Wei, Zhipeng and Guo, Ya and Zhu, Huijia and Chen, Jingjing",
  title = "{EchoingPixels}: Aliasing-Resistant Joint Token Reduction for Audio-Visual LLMs",
  booktitle = ICML,
  year = 2026,
}

@inproceedings{omnisift,
  author = "Ding, Yue and Ji, Yiyan and Li, Jungang and Liu, Xuyang and Chen, Xinlong and Wu, Junfei and Li, Bozhou and Zeng, Bohan and Shi, Yang and Guan, Yushuo and Zhang, Yuanxing and Liu, Jiaheng and Liu, Qiang and Wan, Pengfei and Wang, Liang",
  title = "{OmniSIFT}: Modality-Asymmetric Token Compression for Efficient Omni-Modal Large Language Models",
  booktitle = ICML,
  year = 2026,
}

@inproceedings{objectsthatsound,
  author = "Arandjelovi{\'c}, Relja and Zisserman, Andrew",
  title = "Objects that Sound",
  booktitle = ECCV,
  year = 2018,
}
\clearpage
\makeatletter
\setlength{\@dblfptop}{0pt}
\twocolumn[
\vbox to \titlebox{%
  \hsize\textwidth
  \linewidth\hsize
  \vskip 0.625in minus 0.125in
  \centering
  {\LARGE\bf Supplementary Document\par}%
  \vskip 0.1in plus 0.5fil minus 0.05in
  {\Large\textbf{Deferred Audio Pruning with Local Audio--Visual Dynamics for Omni-LLMs}\par}%
  \vskip .2em plus 0.25fil
  \vskip 1em plus 2fil
}]
\makeatother
\setcounter{section}{0} %
\setcounter{figure}{0}\setcounter{table}{0}\setcounter{equation}{0}\setcounter{algorithm}{0}
\renewcommand{\thefigure}{S\arabic{figure}}
\renewcommand{\thetable}{S\arabic{table}}

\section*{Technical Appendix}
This appendix provides implementation details and extended evidence omitted from the main paper.

The contents are organized as follows:

\noindent\begin{tabularx}{\columnwidth}{@{}lX@{}}
\textbf{Appendix~\ref{sec:implementation}} & Implementation and evaluation details \\
\textbf{Appendix~\ref{sec:additional-efficiency}} & Additional efficiency results \\
\textbf{Appendix~\ref{sec:compression-ablations}} & Additional compression analyses \\
\textbf{Appendix~\ref{sec:audio-evidence}} & Audio preservation and information density \\
\textbf{Appendix~\ref{sec:alignment-scoring}} & Audio--visual alignment scoring \\
\textbf{Appendix~\ref{sec:benchmark-breakdowns}} & Expanded benchmark results \\
\textbf{Appendix~\ref{sec:cost-evaluation}} & Efficiency and cost measurement \\
\textbf{Appendix~\ref{sec:reproducibility}} & Method context
\end{tabularx}\par

\begin{center}
\small
\begin{tabularx}{\columnwidth}{@{}l>{\raggedright\arraybackslash}Xrrl@{}}
\toprule
Benchmark & Evaluation Scope & Videos & QA & Metric \\
\midrule
AVUT & Audio-centric & 698 & 1,734 & Acc. \\
WorldSense & Broad audio--visual & 1,662 & 3,172 & Acc. \\
DailyOmni & Everyday temporal & 684 & 1,196 & Acc. \\
Video-MME & General long-video & 900 & 2,700 & Acc. \\
AVHBench & Hallucination & 2,136 & 5,302 & Acc./F1 \\
\bottomrule
\end{tabularx}
\end{center}
\noindent\textbf{Benchmark summary} AVHBench comprises three binary judgment tasks and additionally reports precision, recall, and the proportion of ``Yes'' predictions.\par

\appendix
\section{Implementation and Evaluation Details}
\label{sec:implementation}

\subsection{Expanded Benchmark Details}
We select benchmarks to test whether compression remains effective across varying audio--visual dependencies,
temporal scales, and reasoning demands. AVUT evaluates audio-centric video understanding through six Audio-Visual Human tasks
designed to reduce reliance on textual shortcuts~\citep{avut}. WorldSense spans 26 tasks across eight real-world domains
and tests coupled audio--visual reasoning~\citep{worldsense}. DailyOmni covers six forms of temporal and
cross-modal reasoning in everyday clips~\citep{dailyomni}, while Video-MME tests general video
understanding across short, medium, and long durations without subtitles~\citep{videomme}. AVHBench is
used separately for three balanced binary tasks that diagnose audio--visual hallucination~\citep{avhbench}.

\begin{algorithm}[!t]
\caption{A-PACK Token Compression}
\label{alg:lace}
\textbf{Input}: audio tokens $X_A$, visual tokens $X_V$, query $Q$\\
\textbf{Parameter}: $w$, $B$, $\tau_{\mathrm{sim}}$, $\tau$, $\gamma$, $\beta$, $L_{\mathrm{mid}}$, $P$\\
\textbf{Output}: compressed token sequence\\
\begin{algorithmic}[1]
\STATE Resample audio embeddings onto the visual-frame time axis and mean-pool each frame.
\STATE Segment at smoothed consecutive-frame cosine $<\tau_{\mathrm{sim}}$.
\STATE Score segment $m$ from frame--query relevance $r_f$ and broadcast the score as $u_f$ to its frames.
\STATE Compute each frame's local-window CKA $c_f$ and the base visual score $b_f=\gamma u_f+(1-\gamma)c_f$.
\STATE \textbf{if} clip-mean CKA $\bar c\le\tau$ \textbf{then} blend $c_f$ with normalized frame--audio distance $\hat d_f$; \textbf{else} set $s_f=b_f$.
\STATE Allocate the remaining video budget $B-|X_A|$ across frames according to $s_f$.
\FOR{each segment}
  \STATE Take the first frame as anchor; the rest as non-anchor.
  \STATE Anchor tokens: density-aware farthest-point sampling (DA-FPS).
  \STATE Non-anchor tokens: keep those least similar to the anchor.
\ENDFOR
\STATE Feed the retained audio and video tokens to the LLM.
\FOR{layer $\ell \ge L_{\text{mid}}$}
  \STATE Keep the top $(1-P)$ fraction by last-query attention; drop the remaining $P$ and slice their KV-cache entries.
\ENDFOR
\STATE \textbf{return} compressed token sequence
\end{algorithmic}
\end{algorithm}

\begin{center}
{\small
\setlength{\tabcolsep}{1.2pt}
\begin{tabular}{@{}lcc@{\hspace{5pt}}lcc@{}}
\toprule
Method & $35\%$ & $25\%$ & Method & $35\%$ & $25\%$ \\
\midrule
Random & T & T & FastV & $.35$ & T \\
DyCoke & T & T & UniComp & $.41$ & $.27$ \\
FlashVID & T & T & OmniSelect & T & T \\
OmniZip & $.40/.70$ & $.75/.75$ & \textbf{A-PACK} & $.50/.10$ & $.35/.15$ \\
\bottomrule
\end{tabular}

\textbf{Budget parameters} $35\%$/$25\%$ settings (T: tuned).
}
\end{center}

\subsection{Algorithm Overview}
Algorithm~\ref{alg:lace} summarizes pre-LLM allocation (Section~3.1) and query-aware inner-LLM pruning (Section~3.2).
\begin{table}[!t]
\centering
\setlength{\tabcolsep}{2pt}
{\small
\begin{tabularx}{\columnwidth}{@{}l>{\raggedright\arraybackslash}Xr@{}}
\toprule
Stage & Hyperparameter & Default \\
\midrule
\multirow{12}{*}{\rotatebox{90}{Pre-LLM}}
 & Continuity threshold $\tau_{\mathrm{sim}}$ & $0.97$ \\
 & Continuity smoothing $\sigma$        & $1.0$ \\
 & Segment weight                       & $(\max r+\operatorname{mean}r)\sqrt{\mathrm{dur}}$ \\
 & DA-FPS density $k$                   & $5$ \\
 & Anchor priority                     & $0.3$ \\
 & Window size $w$                     & $3$ \\
 & Visual-score weight $\gamma$           & $0.9$ \\
 & Dedup coverage threshold            & $0.8$ \\
 & DA-FPS power                        & $0.5$ \\
 & CKA gate $\tau$                     & $0.78$ \\
 & Fine-blend $\beta$                 & $0.90$ \\
 & Pre-LLM retention $R_{\mathrm{pre}}$ (35\%/25\%)          & $0.50/0.35$ \\
\midrule
\multirow{3}{*}{\rotatebox{90}{Inner}}
 & Mid layer $L_{\mathrm{mid}}$        & depth$/2$ ($14$/$18$ on 7B/3B) \\
 & Per-layer pruning $P$ (35\%/25\%)    & $10\%/15\%$ \\
 & Query score                         & last-query only \\
\bottomrule
\end{tabularx}%
}
\caption{A-PACK hyperparameters used in main evaluations. Paired values correspond to the $35\%$/$25\%$ prefill-FLOPs tiers; $R_{\mathrm{pre}}$ and $P$ denote pre-LLM retention and per-layer pruning, respectively.}
\label{tab:hparams}

\medskip
\centering
\setlength{\tabcolsep}{2.5pt}
\renewcommand{\arraystretch}{1.0}
{\small
\begin{tabularx}{\columnwidth}{@{}l*{4}{>{\raggedleft\arraybackslash}X}@{}}
\toprule
\multirow[t]{2}{*}{Method} & \multicolumn{2}{c}{Resource and Quality} &
\multicolumn{2}{c}{Inference Speed} \\
\cmidrule(lr){2-3}\cmidrule(lr){4-5}
& \shortstack{Memory\\(GB)$\downarrow$} &
\shortstack{Accuracy\\(\%)$\uparrow$} &
\shortstack{Prefill\\Speedup$\uparrow$} &
\shortstack{End-to-end\\Speedup$\uparrow$} \\
\midrule
Full & 25.1 & 46.8 & 1.00$\times$ & 1.00$\times$ \\
FlashVID & 25.5 & 43.5 & 1.60$\times$ & 1.59$\times$ \\
Random & 25.5 & \underline{43.7} & 1.74$\times$ & 1.72$\times$ \\
DyCoke & 25.6 & 40.3 & 1.65$\times$ & 1.65$\times$ \\
FastV & 25.7 & 41.0 & 1.68$\times$ & 1.67$\times$ \\
UniComp & 25.5 & 42.1 & 1.63$\times$ & 1.61$\times$ \\
OmniZip & \underline{22.2} & 42.7 & \underline{1.83}$\times$ & \underline{1.83}$\times$ \\
\rowcolor{mycol1}
\textbf{A-PACK} & \textbf{21.9} & \textbf{45.3} & \textbf{1.94}$\times$ & \textbf{1.92}$\times$ \\
\bottomrule
\end{tabularx}%
}
\caption{Efficiency on WorldSense for Qwen2.5-Omni-7B in the $25\%$ prefill-FLOPs tier. Best and second-best compressor results are bold and underlined, respectively.}
\label{tab:eff-ws}
\end{table}

Here, $B$ is the joint audio--video pre-LLM budget, $L_{\mathrm{mid}}$ is the first decoder pruning layer, and $P$ is the fraction removed at each such layer. We use $P{=}10\%$ and $15\%$ for the $35\%$ and $25\%$ tiers, respectively; Table~\ref{tab:hparams} defines the remaining controls.

\subsection{Compression Protocol and Hyperparameters}
All experiments use the public Qwen2.5-Omni 7B/3B checkpoints without fine-tuning. We define the
$35\%$ and $25\%$ prefill-FLOPs tiers relative to Full Tokens, rather than by nominal token
retention. We match each tier by varying the method-specific parameters summarized below.

\emph{Tuned} denotes method-specific values calibrated separately for each backbone and benchmark. FastV applies inner-LLM pruning
  and retains its released $0.35$ setting at the $35\%$ tier, whereas UniComp compresses only visual tokens and therefore uses a
  relatively higher visual retention to match the FLOPs tiers. For OmniZip and A-PACK, each slash-separated pair gives the audio/
  video settings, respectively. Random, DyCoke, FlashVID, and OmniSelect use their native parameters tuned for each tier; Random uses
  seed~$0$. Unless stated otherwise, A-PACK uses Table~\ref{tab:hparams}.

\begin{figure}[!t]
\centering
\includegraphics[width=0.92\columnwidth]{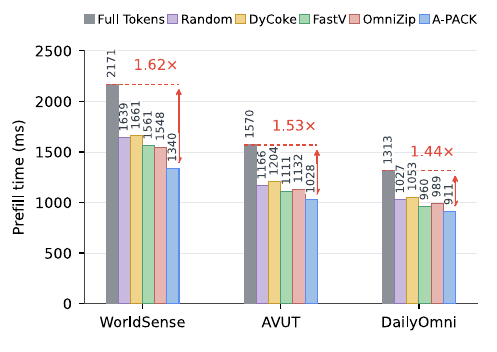}
\caption{Prefilling time on Qwen2.5-Omni-3B across three benchmarks.
A-PACK is the fastest.}
\label{fig:prefill3b}
\end{figure}

\begin{figure*}[!t]
\centering
{\setlength{\tabcolsep}{1pt}
\begin{tabular}{@{}cc@{}}
\includegraphics[width=0.574\textwidth]{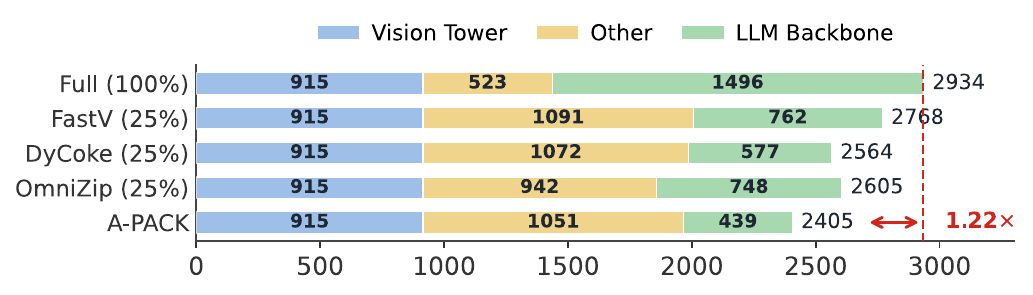} &
\includegraphics[width=0.406\textwidth]{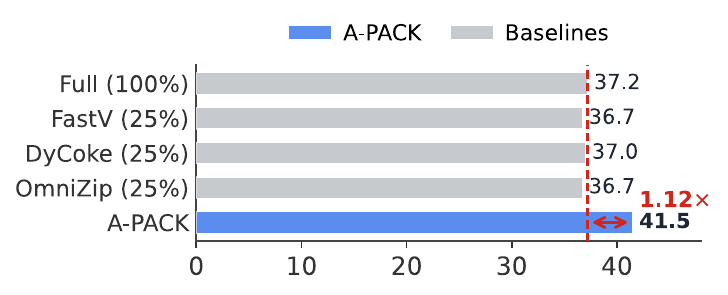} \\
\textbf{(a)} TTFT breakdown & \textbf{(b)} Decoding throughput
\end{tabular}}
\caption{Inference speed on Qwen2.5-Omni-3B at the $25\%$ prefill-FLOPs tier.
A-PACK achieves the shortest time-to-first-token and highest decoding throughput.}
\label{fig:infspeed3b}
\end{figure*}

\begin{figure}[!t]
\centering
\includegraphics[width=\columnwidth]{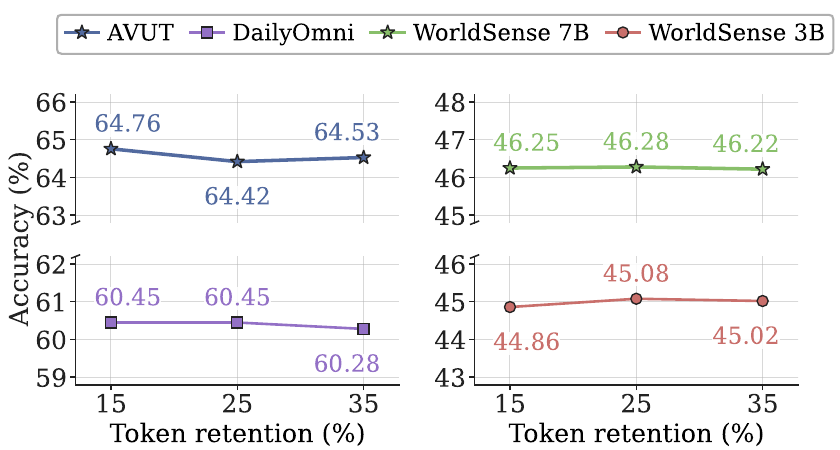}
\caption{Accuracy versus final token retention. \textbf{Left:} AVUT and DailyOmni on 7B.
\textbf{Right:} WorldSense on 7B and 3B.}
\label{fig:targetret}
\end{figure}

\subsection{Pre-LLM Compression Details}
This section details the pre-LLM stage in Section~3.1. The visual-score weight $\gamma$ balances query-relevant segment importance against the local CKA score when ranking frames. We first mean-pool the backbone tokens within each frame. After Gaussian smoothing (width $\sigma$), a drop in cosine similarity between two consecutive
frame embeddings below $\tau_{\mathrm{sim}}$ starts a new scene segment. We then distribute the visual-token
budget across the resulting segments: segments more relevant to the query receive more tokens, with a
sublinear duration weight so that long segments do not dominate. Anchor tokens (the first frame of
each segment) are selected by \emph{density-aware farthest-point sampling} (DA-FPS). Starting from the
densest prototype, we iteratively add the token that maximizes
$(1-\text{sim}_{\max}^{\text{sel}})\cdot\rho$, where $\text{sim}_{\max}^{\text{sel}}$ is the maximum
cosine similarity to already-selected tokens and the density $\rho$ is a token's mean top-$k$ cosine
similarity weighted by how many tokens it covers. DA-FPS thus rewards both density and novelty, keeping a
\emph{diverse, high-coverage} set of representatives rather than clustering around a few peaks.
Non-anchor frames keep only the tokens least similar to their paired anchor (deduplication), dropping
content the anchor already represents. Audio tokens are not pruned in this stage. The pre-LLM budget
counts audio and video tokens together: if the total budget is $B$ and all $N_A$ audio tokens are kept,
we allocate the remaining $B-N_A$ tokens to video. The video keep ratio is then chosen to retain exactly
this number of video tokens.

\section{Additional Efficiency Results}
\label{sec:additional-efficiency}
The $3$B backbone reproduces the $7$B inference-speed ordering: A-PACK has the lowest prefilling time (Figure~\ref{fig:prefill3b}), the shortest time-to-first-token, and the highest decoding throughput (Figure~\ref{fig:infspeed3b}). Table~\ref{tab:eff-ws} further shows the best compressed-model accuracy together with the lowest memory and largest speedups on WorldSense.

\begin{table}[!t]
\centering
\renewcommand{\arraystretch}{1.0}
\setlength{\tabcolsep}{2pt}
{\small
\begin{tabularx}{\columnwidth}{@{}*{5}{>{\centering\arraybackslash}X}@{}}
\toprule
\multicolumn{2}{c}{Method} & \multicolumn{2}{c}{Benchmark} &
\multirow[t]{2}{*}{FLOPs$\downarrow$} \\
\cmidrule(lr){1-2}\cmidrule(lr){3-4}
Pre-LLM & Inner-LLM & AVUT (\%) & DailyOmni (\%) & \\
\midrule
Full & Full & 62.5 & 61.6 & 100\% \\
Random & Random & 57.3 & 55.0 & 35.8\% \\
Random & A-PACK & 57.7 & 53.8 & \textbf{28.8}\% \\
A-PACK & Random & 59.0 & \textbf{56.2} & 35.8\% \\
A-PACK & FastV & 50.6 & 55.7 & \underline{30.0}\% \\
A-PACK & DyCoke & \underline{59.6} & \underline{56.1} & 37.3\% \\
\rowcolor{mycol1}
\textbf{A-PACK} & \textbf{A-PACK} & \textbf{60.6} & 55.4 & \textbf{28.8}\% \\
\bottomrule
\end{tabularx}%
}
\caption{Stage-combination ablation on Qwen2.5-Omni-3B (columns as in Table~\ref{tab:stagecombo-7b}).}
\label{tab:stagecombo-3b}
\end{table}
\begin{table}[!t]
\centering
\renewcommand{\arraystretch}{1.0}
\setlength{\tabcolsep}{2pt}
{\small
\begin{tabularx}{\columnwidth}{@{}lcc*{5}{>{\centering\arraybackslash}X}@{}}
\toprule
\multirow[t]{2}{*}{Method} & \multicolumn{2}{c}{Stage} &
\multicolumn{1}{c}{Cost} & \multicolumn{4}{c}{Benchmark} \\
\cmidrule(lr){2-3}\cmidrule(lr){4-4}\cmidrule(lr){5-8}
& \shortstack{Pre-\\LLM} & \shortstack{Inner-\\LLM} &
\shortstack{Prefill\\FLOPs} & AVUT & \shortstack{World\\Sense} &
\shortstack{Daily\\Omni} & Avg. \\
\midrule
\multicolumn{8}{c}{\emph{Qwen2.5-Omni-7B}} \\
\midrule
Full     & -- & -- & 100\% & 64.5 & 46.8 & 63.0 & 100\% \\
Pre only & \checkmark & -- & 44.6\% & 64.5 & 46.2 & 59.4 & 97.7\% \\
\rowcolor{mycol1}
\textbf{A-PACK} & \checkmark & \checkmark & \textbf{29.9\%} & 64.2 & 45.8 & 60.5 & \textbf{97.8\%} \\
\midrule
\multicolumn{8}{c}{\emph{Qwen2.5-Omni-3B}} \\
\midrule
Full     & -- & -- & 100\% & 62.5 & 46.2 & 61.6 & 100\% \\
Pre only & \checkmark & -- & 41.8\% & 60.5 & 45.3 & 56.8 & 95.7\% \\
\rowcolor{mycol1}
\textbf{A-PACK} & \checkmark & \checkmark & \textbf{23.1\%} & 60.6 & 45.3 & 57.8 & \textbf{96.3\%} \\
\bottomrule
\end{tabularx}%
}
\caption{Sequential stage ablation: \emph{Pre only} keeps the pre-LLM stage, and \emph{A-PACK} adds the
inner-LLM stage. Prefill FLOPs are relative ratio versus Full tokens. \emph{Avg.} is the mean of
AVUT/WorldSense/Daily(Omni) relative to Full.}
\label{tab:stageseq}
\end{table}

\begin{figure*}[t]
\centering
\begin{subfigure}[t]{0.31\textwidth}
\centering
\includegraphics[width=\linewidth]{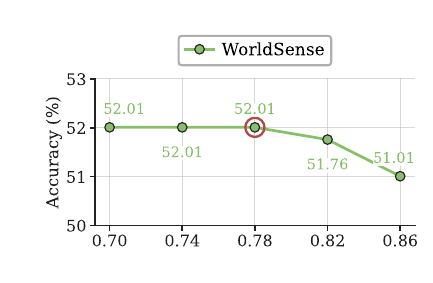}
\caption{Gate-threshold sensitivity}
\label{fig:hparam-gate}
\end{subfigure}\hfill
\begin{subfigure}[t]{0.31\textwidth}
\centering
\includegraphics[width=\linewidth]{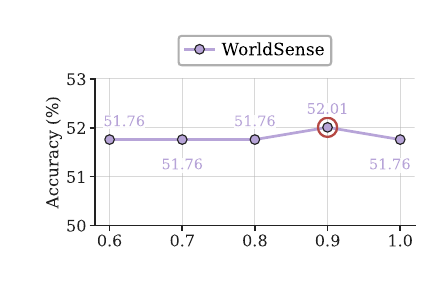}
\caption{Fine-blend sensitivity}
\label{fig:hparam-blend}
\end{subfigure}\hfill
\begin{subfigure}[t]{0.31\textwidth}
\centering
\includegraphics[width=\linewidth]{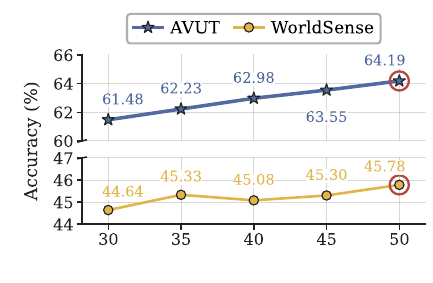}
\caption{Pre-LLM retention ratio (\%)}
\label{fig:preratio}
\end{subfigure}
\caption{Additional sensitivity analyses. Red circles denote
the highest scores.}
\label{fig:additional-sweeps}
\end{figure*}
\begin{table*}[!t]
\centering
\setlength{\tabcolsep}{2pt}
{\small
\begin{tabularx}{\textwidth}{@{}>{\raggedright\arraybackslash}p{0.14\textwidth}
>{\raggedright\arraybackslash}p{0.25\textwidth}
>{\raggedright\arraybackslash}X >{\raggedright\arraybackslash}X@{}}
\toprule
Component & Axis (selected) & Tested value(s) & $\Delta$ accuracy (points) \\
\midrule
\multirow{2}{*}{Segmentation}
& Continuity $\tau_{\mathrm{sim}}$ ($0.97$) & $0.93,\ 0.95,\ 0.99$ & $+1.76,\ +0.75,\ -2.51$ \\
& Smoothing $\sigma$ ($1.0$) & $0.5,\ 1.5,\ 2.0$ & $0.00,\ 0.00,\ -1.26$ \\
\midrule
\multirow{4}{0.14\textwidth}{Token selection}
& DA-FPS power ($0.5$) & $0.3,\ 0.7,\ 0.9$ & $+1.26,\ -1.51,\ -1.76$ \\
& DA-FPS density $k$ ($5$) & $3,\ 7,\ 9,\ 11$ & $+0.75,\ +0.75,\ -0.50,\ -0.50$ \\
& Dedup coverage ($0.8$) & $0.7,\ 0.9,\ 1.0$ & $-1.51,\ -1.51,\ -0.75$ \\
& Anchor priority ($0.3$) & $0.2,\ 0.4,\ 0.5$ & $0.00,\ -1.51,\ +1.26$ \\
\midrule
\multirow{4}{*}{Alignment}
& Temporal window $w$ ($3$) & $5,\ 7,\ 9$ & $-3.52,\ -3.77,\ -2.26$ \\
& Spatial CKA & $w{=}1$ & $-4.02$ \\
& Gate $\tau$ ($0.78$) & $0.70,\ 0.74,\ 0.82,\ 0.86$ & $0.00,\ 0.00,\ -0.25,\ -1.01$ \\
& Fine-blend $\beta$ ($0.9$) & $0.6,\ 0.7,\ 0.8,\ 1.0$ & $-0.25,\ -0.25,\ -0.25,\ -0.25$ \\
\bottomrule
\end{tabularx}
}
\caption{One-at-a-time parameter sensitivity on a WorldSense subset.
Each cell reports the accuracy change $\Delta$ from the selected setting shown in parentheses.}
\label{tab:ablation}
\end{table*}
\section{Additional Compression Analyses}
\label{sec:compression-ablations}
\subsection{Stage Contributions}
The $3$B backbone reproduces the $7$B stage-combination finding
(Table~\ref{tab:stagecombo-3b}). All non-base rows use a $50\%$ pre-LLM retention ratio
and per-layer $P{=}15\%$. At the lowest prefill cost, using A-PACK in
both stages gives the highest AVUT accuracy and the strongest overall trade-off. Although
random inner pruning is $0.8$ points higher on DailyOmni, it requires about $24\%$ more prefill FLOPs
and is therefore not comparable at the same cost.

Adding the stages one at a time (Table~\ref{tab:stageseq}) isolates each
contribution. The pre-LLM stage alone already retains $97.7\%$ of full accuracy at
$44.6\%$ prefill on 7B, and adding the inner-LLM stage reduces prefill from $44.6\%$ to $29.9\%$ while holding
$97.8\%$. The 3B backbone behaves the same.

\subsection{Retention and Parameter Sensitivity}
\paragraph{Total token budget.} Figure~\ref{fig:targetret} varies the proportion of original
audio--visual tokens retained after both stages. The pre-LLM stage first keeps about $50\%$, and
inner-LLM pruning reduces the sequence to $15\%$, $25\%$, or $35\%$. Accuracy remains stable,
changing by at most $0.4$ points on AVUT, $0.2$ on DailyOmni, and $0.2$ on WorldSense. This stability
indicates that inner-LLM compression preserves performance across retention targets, even at $15\%$.

\paragraph{Additional sensitivity analyses.}
Figure~\ref{fig:additional-sweeps} first examines the gate threshold, fine-blend
weight, and pre-LLM retention ratio. The gate has a large local effect, the
fine-blend weight is comparatively stable, and increased pre-LLM retention yields
the expected accuracy--cost trade-off.

\paragraph{Individual parameter sensitivity.} Table~\ref{tab:ablation} shows that the
continuity threshold is the most sensitive segmentation control: changing it moves scene
boundaries and consequently changes the budget and selection of every segment. Since its
best value varies across benchmarks, we use $\tau_{\mathrm{sim}}{=}0.97$ as a robust shared
setting. Lower DA-FPS powers perform better because they temper the density term and retain
novelty-driven coverage, whereas large powers overemphasize dense token neighborhoods.
Likewise, $k{=}9$ and $11$ average density over overly broad neighborhoods, blurring the
local structure needed for a diverse anchor set. Gate and fine-blend settings are comparatively
stable near the selected values. Consistent with Figure~\ref{fig:hparamsens}, temporal CKA
peaks at $w{=}3$; wider windows and the spatial-CKA baseline are worse.

\section{Audio Utility Analysis}
\label{sec:audio-evidence}
\subsection{Per-Token Information Density}
\paragraph{Measurement.} This is the protocol behind Figure~\ref{fig:density}. For each
modality $m\in\{A,V\}$, we characterize task-relevant information density as
\begin{equation}
\eta_m = \frac{I(Y;X_m\mid Q,S_0)}{C_m},
\label{eq:task-density}
\end{equation}
where $Y$ is the answer, $X_m$ the modality tokens, $Q$ the query, $S_0$ the remaining
input, and $C_m$ the token cost. Over an interval of duration $\Delta$, $C_m=R_m\Delta$.
Audio is represented at about $R_A{\approx}25$ tokens per second, far below the visual rate
$R_V$; preserving it therefore consumes little of the budget while retaining compact evidence.
We estimate task-relevant information density by mean-replacement occlusion. We store the
answer distribution $p_{\text{base}}=\mathrm{softmax}(\text{logits})$, replace block $k$ with
its mean embedding, and re-run only the decoder with the sequence length and positional
encodings fixed. For fair comparison, a block is one video frame and the audio tokens span the
same time width. We report the per-token normalized divergence as follows.
\begin{equation}
\text{KL/token}_m=\frac{1}{|\mathcal{B}_m|}\sum_{k\in\mathcal{B}_m}
\frac{\mathrm{KL}\!\big(p_{\text{base}}\,\|\,p_{\text{occ}}^{(k)}\big)}{|\text{block}_k|}
\label{eq:kltoken}
\end{equation}
We measure representational diversity per token using effective rank~\citep{erank}:
\begin{equation}
\mathrm{erank}(X)=\exp\!\big(H(\tilde\sigma)\big),
\qquad H(\tilde\sigma)=-\sum_i\tilde\sigma_i\ln\tilde\sigma_i,
\label{eq:effective-rank}
\end{equation}
where $H$ is the Shannon entropy of the normalized singular-value spectrum $\tilde\sigma$.
Effective rank counts the independent directions used by a token set, so a lower value indicates
greater redundancy. Both measurements use AVUT and WorldSense.

\paragraph{Statistical test.} For each metric, we compare paired audio and video values from
the same AVUT and WorldSense evaluation sample using a one-sided Wilcoxon signed-rank test with the
alternative that audio exceeds video. Both task-relevant information density and representational
diversity are significantly higher for audio. The test evaluates the consistency of the audio--video
difference across samples rather than treating individual tokens as independent observations.

\subsection{Audio Preservation under a Fixed Budget}
\begin{figure}[!t]
\centering
\includegraphics[width=0.98\columnwidth]{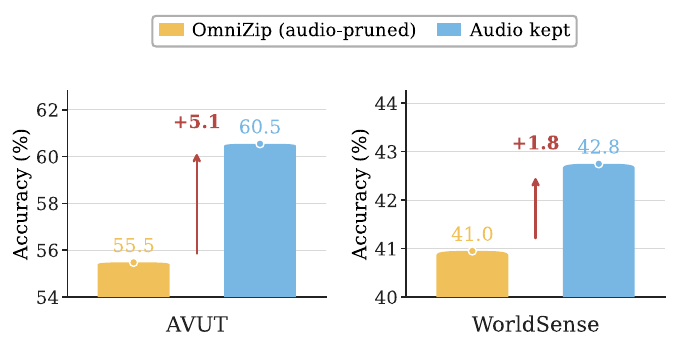}
\caption{Audio guidance versus audio pruning at the $25\%$ budget. Keeping audio as guidance
improves accuracy on both AVUT and WorldSense.}
\label{fig:nap}
\end{figure}

Figure~\ref{fig:nap} compares OmniZip, which prunes audio, against a not-audio-pruned (NAP) variant that
keeps audio purely as guidance under the same overall budget. NAP disables OmniZip's audio-token
merge and masking operation, keeps every audio position, and adjusts video merging to preserve the same
total budget. Removing audio pruning improves accuracy
on both benchmarks, with the largest gain on AVUT ($+5.1$) and a consistent gain on WorldSense ($+1.8$).
Under a fixed budget, discarding audio therefore loses more evidence than the extra visual tokens recover.

\begin{figure}[!t]
\centering
\includegraphics[width=0.985\columnwidth]{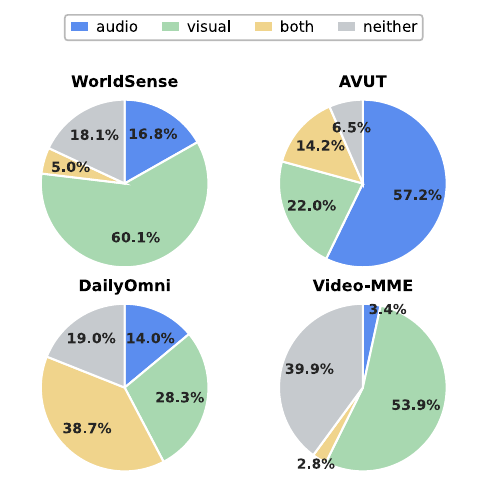}
\caption{Question-modality categories assigned by an LLM judge on WorldSense,
AVUT, DailyOmni, and Video-MME.}
\label{fig:querymod}
\end{figure}

\begin{figure}[!t]
\centering
\includegraphics[width=\columnwidth]{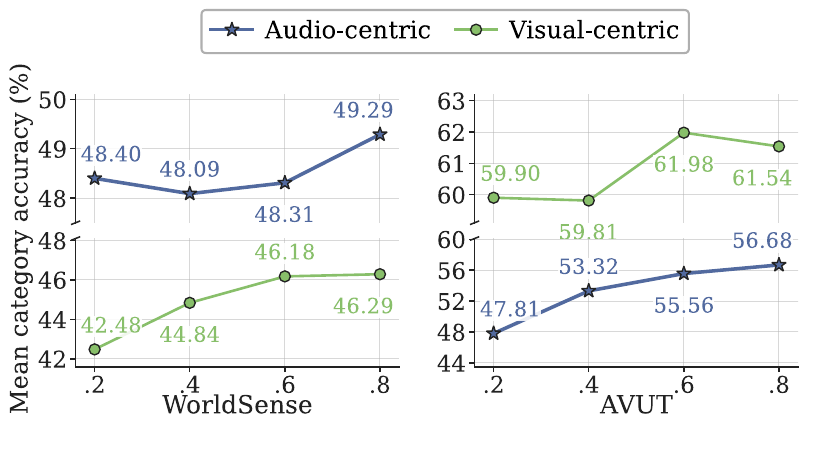}
\caption{Per-category audio-keep sweep, decomposing Figure~\ref{fig:audiopruning} by the query-modality buckets
of Figure~\ref{fig:querymod} (mean category accuracy vs.\ audio keep ratio; WorldSense $6$ audio-/$20$
visual-centric categories, AVUT $4$/$2$).}
\label{fig:audiocentric}
\end{figure}

\paragraph{Which modality answers, and where audio matters.} An LLM judge classifies the modality
indicated by each question and its answer choices as audio, visual, both, or neither
(Figure~\ref{fig:querymod}). AVUT is predominantly audio-oriented, whereas WorldSense is
unexpectedly visual-oriented. DailyOmni is comparatively balanced across audio and visual
requirements, making it well suited to omni-modal evaluation, whereas Video-MME is predominantly
visual-centric. Figure~\ref{fig:audiocentric} then decomposes the audio-keep
sweep of Figure~\ref{fig:audiopruning} by these same buckets. On AVUT the audio-centric categories gain $+8.9$
points as the audio-keep ratio grows from $0.2$ to $0.8$, while the visual-centric categories stay high
and nearly flat. On WorldSense both buckets improve as audio is kept. Audio is thus most valuable
precisely where the question requires it, supporting its use as guidance rather than an early pruning target.

\section{Audio--Visual Alignment Scoring}
\label{sec:alignment-scoring}

\subsection{Alternative Alignment Scorers}
\paragraph{Alignment-scoring alternatives.} Table~\ref{tab:alignment} compares alignment-scoring functions while keeping the remaining compressor fixed. \emph{Base} combines the native visual-relevance score with local CKA. \emph{Random} uses deterministic pseudo-random frame scores that are independent of the audio, video, and question content. \emph{Attention} scores frames from query--video attention computed with the decoder's query and key projections. \emph{Audio-saliency} promotes frames co-timed with audio tokens that differ most from their local audio context. \emph{AV cosine} compares every visual token with every audio token, assigns each visual token its strongest cross-modal cosine match, and averages the scores within a frame. \emph{CKA} measures the similarity of audio and visual representation structure in a short time-aligned window. \emph{OT} assigns frame scores through a global cost-aware matching of all audio and visual tokens. \emph{CKA$+$OT} blends normalized CKA and OT scores. \emph{Gated CKA} uses the visual--CKA score for strongly synchronized clips and otherwise combines local CKA with a token-level audio--visual distance cue.
\paragraph{Per-task AVHBench breakdown.} Table~\ref{tab:alignment} pools the three AVHBench
yes/no judgment tasks. Table~\ref{tab:avhbench-tasks} separates the same scoring-rule ablation into
audio-driven video hallucination (A$\to$V), video-driven audio hallucination (V$\to$A), and
audio--visual matching (Match)~\citep{avhbench}. Gated CKA (A-PACK) most improves the two
alignment-sensitive tasks (A$\to$V and Match).
\begin{table}[!t]
\centering
{\small
\begin{tabularx}{\columnwidth}{@{}l*{3}{>{\centering\arraybackslash}X}@{}}
\toprule
Scoring rule & A$\to$V & V$\to$A & Match \\
 & Acc.$\uparrow$ & Acc.$\uparrow$ & Acc.$\uparrow$ \\
\midrule
Base frame score & 72.5 & \underline{76.2} & \underline{71.1} \\
Random score & 73.7 & \underline{76.2} & 70.3 \\
Attention score & 73.5 & \underline{76.2} & 70.0 \\
Audio-saliency score & 73.9 & 75.9 & 70.3 \\
Audio-video cosine & \underline{74.5} & \textbf{76.3} & 69.9 \\
Optimal Transport & 73.5 & 76.0 & 70.0 \\
CKA $+$ OT & 73.6 & 75.9 & 69.8 \\
\rowcolor{mycol1}
\textbf{Gated CKA (A-PACK)} & \textbf{74.7} & 75.0 & \textbf{72.5} \\
\bottomrule
\end{tabularx}%
}
\caption{Per-task AVHBench accuracy for alignment-scoring rules. Best and second-best
results in each column are bold and underlined, respectively.}
\label{tab:avhbench-tasks}
\end{table}

\subsection{Paired Prediction Analysis}
The matched WorldSense comparison in the main paper changes only the alignment scorer and evaluates
on WorldSense. It compares Gated CKA and static cosine on identical questions; an exact two-sided
McNemar test evaluates whether their discordant paired predictions favor one scorer. Together with the
domain-level gains, the significant result supports local temporal correspondence as a useful selection
signal.

\section{Expanded Benchmark Results}
\label{sec:benchmark-breakdowns}
This section expands the consolidated 7B/3B main table (Table~\ref{tab:main7b}) with full per-benchmark
evidence. Tables~\ref{tab:avut} and~\ref{tab:worldsense} give the per-category AVUT (with Video-MME) and
per-domain WorldSense breakdowns for both backbones. These carry the detailed cost columns
(prefilling FLOPs, FLOPs ratio, and before-LLM retention) defined in Appendix~G.
A-PACK stays closest to full-token accuracy on average across categories and domains
in both prefill-FLOPs tiers.

\newcommand{\ExpandedBenchmarkTables}{%
\begin{table*}[!t]
\centering
\renewcommand{\arraystretch}{1.0}
\setlength{\tabcolsep}{3.8pt}
{\small
\begin{tabular}{lc*{11}{r}}
\toprule
\multirow[t]{2}{*}{Method} & \multirow[t]{2}{*}{$R$ (\%)} & \multicolumn{3}{c}{Cost} & \multicolumn{7}{c}{AVUT-Bench} & \multicolumn{1}{c}{Video-MME} \\
\cmidrule(lr){3-5}\cmidrule(lr){6-12}\cmidrule(lr){13-13}
 & & FLOPs (T) $\downarrow$ & Ratio $\downarrow$ & Pre-LLM $\downarrow$ & EL $\uparrow$ & OR $\uparrow$ & OM $\uparrow$ & IE $\uparrow$ & CC $\uparrow$ & CM $\uparrow$ & Avg. $\uparrow$ & Avg. $\uparrow$ \\
\midrule
\multicolumn{13}{c}{\emph{Qwen2.5-Omni-7B}} \\
\midrule
\rowcolor{mycol2!45!white}
Full Tokens & 100 & 62.2 & 100.0\% & 100\% & 38.2 & 67.8 & 59.6 & 85.6 & 44.1 & 66.7 & 64.5 & 66.0 \\
\midrule
Random & 35 & \underline{21.3} & 35.4\% & \underline{40.0\%} & 37.6 & 66.2 & 54.8 & 78.8 & 39.8 & 63.3 & 60.7 & 65.6 \\ 
DyCoke (V\&A)~\citeyearpar{dycoke} & 35 & \textbf{21.1} & 35.1\% & \textbf{39.7\%} &\textbf{44.1}& 64.6 & 56.1 & 71.5 & 35.6 & 64.3 & 59.9 & 66.0 \\ 
FlashVID~\citeyearpar{flashvid} & 35 & 21.5 & \underline{34.6\%} & 44.7\% & 38.8 & 60.8 & 54.1 & 83.1 & 40.7 & 65.2 & 61.0 & OOM \\ 
FastV~\citeyearpar{fastv} & 35 & 23.9 & 38.4\% & 100\% & 38.8 & 66.9 & 54.1 & 84.4 &\textbf{44.1}& 55.4 & 60.2 & OOM        \\ 
UniComp~\citeyearpar{unicomp} & 35 & 21.8 & 36.1\% & 40.8\% &\underline{39.4}& 67.5 & 53.1 &\underline{85.6}&\underline{43.2}& 60.4 & 61.5 & 65.9 \\ 
OmniSelect~\citeyearpar{omniselect} & 35 & 24.1 & 38.7\% & 43.5\% & 34.1 &\textbf{72.0}&\underline{56.9}& 80.1 & 39.0 &\textbf{66.9}&\underline{62.9}& 63.6 \\ 
OmniZip~\citeyearpar{omnizip} & 35 & 21.5 & \textbf{34.5\%} & 40.2\% & 35.9 & 67.5 & 54.8 & 84.7 & 40.7 & 65.2 & 62.4 &\underline{66.6} \\
\rowcolor{mycol1}
\textbf{A-PACK} & 35 & 21.6 & 34.7\% & 51.0\% & 36.5 &\underline{68.8}&\textbf{57.7}&\textbf{86.2}&\underline{43.2}&\underline{65.9}&\textbf{64.0}&\textbf{66.7} \\
\midrule
Random & 25 & 14.8 & 25.0\% & \underline{28.5\%} &\underline{38.8}& 63.3 & 52.0 & 71.8 & 38.1 & 62.1 & 58.0 & 65.6 \\ 
DyCoke (V\&A)~\citeyearpar{dycoke} & 25 & 16.6 & 28.0\% & 31.9\% &\textbf{42.4}& 64.0 &\underline{55.4}& 67.5 & 31.4 &\textbf{63.5}& 58.3 & 66.1 \\ 
FlashVID~\citeyearpar{flashvid} & 25 & 15.7 & 25.1\% & 31.5\% &\underline{38.8}& 58.2 & 53.8 & 85.0 &\underline{39.8}&\underline{62.8}&\underline{60.2}& OOM \\ 
FastV~\citeyearpar{fastv} & 25 & 16.3 & 26.1\% & 100.0\% &\underline{38.8}& 55.0 & 43.9 &\textbf{86.5}&\underline{39.8}& 39.6 & 52.1 & 64.0         \\ 
UniComp~\citeyearpar{unicomp} & 25 & \underline{14.3} & \underline{23.7\%} & \textbf{27.1\%} &\underline{38.8}& 64.6 & 50.3 & 85.3 &\underline{39.8}& 47.5 & 56.9 &\textbf{66.4} \\
OmniSelect~\citeyearpar{omniselect} & 25 & 18.9 & 30.4\% & 34.7\% & 36.5 & 65.3 & 53.1 & 73.3 & 35.6 & 60.2 & 58.0 & 65.9 \\ 
OmniZip~\citeyearpar{omnizip} & 25 & 16.2 & 26.0\% & 31.0\% & 35.9 &\textbf{67.5}& 54.1 & 73.3 & 34.7 &\textbf{63.5}& 59.3 & 65.6         \\ 
\rowcolor{mycol1}
A-PACK & 25 & \textbf{13.9} & \textbf{22.4\%} & 36.5\% & 34.7 &\underline{67.2}&\textbf{56.6}&\underline{85.6}&\textbf{42.4}& 62.4 &\textbf{62.2}&\underline{66.2} \\
\midrule
\multicolumn{13}{c}{\emph{Qwen2.5-Omni-3B}} \\
\midrule
\rowcolor{mycol2!45!white}
Full Tokens & 100 & 31.8 & 100.0\% & 100\% & 32.9 & 65.3 & 58.4 & 85.0 & 44.1 & 62.6 & 62.5 & 62.6 \\
\midrule
Random & 35 & 10.6 & 35.0\% & 42.3\% &\underline{32.4}& 58.8 &\textbf{55.9}& 79.4 & 43.2 & 60.2 & 58.7 & 61.8 \\ 
DyCoke (V\&A)~\citeyearpar{dycoke} & 35 & 10.6 & 34.9\% & 42.3\% & 31.8 & 61.7 & 54.1 & 74.2 & 40.7 & 59.5 & 57.4 &\textbf{62.9} \\
FlashVID~\citeyearpar{flashvid} & 35 & 11.1 & 35.3\% & 53.3\% & 29.4 & 62.1 &\underline{55.6}&\underline{85.3}& 43.2 & 60.7 &\underline{60.1}& OOM \\ 
FastV~\citeyearpar{fastv} & 35 & 10.9 & 34.4\% & 100\% &\underline{32.4}& 60.8 & 47.7 & 84.7 & 40.7 & 53.5 & 56.4 & OOM        \\ 
UniComp~\citeyearpar{unicomp} & 35 & 10.2 & 33.5\% & \underline{40.7\%} & 29.4 &\underline{63.0}& 53.8 &\textbf{85.6}&\textbf{44.9}& 59.7 & 59.9 & 62.4 \\ 
OmniSelect~\citeyearpar{omniselect} & 35 & 11.2 & 35.1\% & 41.9\% &\textbf{35.3}& 62.1 & 51.8 & 82.8 & 43.2 & 56.8 & 58.5 &\underline{62.6} \\
OmniZip~\citeyearpar{omnizip} & 35 & \underline{10.1} & \underline{31.8\%} & \textbf{40.0\%} & 27.6 &\textbf{63.3}& 53.8 & 84.7 & 42.4 &\underline{62.6}&\underline{60.1}& 62.4         \\ 
\rowcolor{mycol1}
A-PACK & 35 & \textbf{9.8} & \textbf{30.8\%} & 51.0\% & 28.9 &\textbf{63.3}& 54.5 & 84.8 &\underline{43.5}&\textbf{63.3}&\textbf{60.7}& 62.5         \\ 
\midrule
Random & 25 & 7.5 & 25.0\% & 31.2\% &\underline{32.9}& 58.2 &\textbf{54.3}& 72.7 & 40.7 & 59.2 & 56.6 & 61.5 \\ 
DyCoke (V\&A)~\citeyearpar{dycoke} & 25 & 7.4 & 25.0\% & 31.2\% & 30.6 & 56.3 & 53.1 & 61.0 & 37.3 & 58.8 & 53.2 & 61.6 \\ 
FlashVID~\citeyearpar{flashvid} & 25 & 8.0 & 25.4\% & 37.8\% & 31.8 & 58.8 &\underline{53.6}&\underline{85.0}&\textbf{44.9}&\underline{59.7}&\underline{59.2}& OOM \\ 
FastV~\citeyearpar{fastv} & 25 & 7.8 & 25.0\% & 100.0\% & 28.8 & 53.1 & 43.6 &\underline{85.0}& 35.6 & 43.2 & 51.0 & 60.1         \\ 
UniComp~\citeyearpar{unicomp} & 25 & \underline{6.5} & \underline{21.6\%} & \textbf{27.1\%} & 29.4 & 60.1 & 47.7 &\textbf{85.3}& 39.8 & 47.2 & 54.6 & 60.4 \\ 
OmniSelect~\citeyearpar{omniselect} & 25 & 8.0 & 25.1\% & 31.1\% &\textbf{35.9}& 56.9 & 49.0 & 75.5 & 41.5 & 55.2 & 55.1 & 61.6 \\ 
OmniZip~\citeyearpar{omnizip} & 25 & 7.5 & 23.7\% & \underline{30.9\%} & 28.8 &\underline{60.8}& 50.8 & 76.7 &\underline{43.2}&\textbf{60.2}& 57.0 &\underline{61.8} \\
\rowcolor{mycol1}
A-PACK & 25 & \textbf{6.2} & \textbf{19.4\%} & 36.5\% & 27.6 &\textbf{63.3}&\textbf{54.3}&\underline{85.0}& 40.7 & 59.0 &\textbf{59.3}&\textbf{61.9} \\
\bottomrule
\end{tabular}%
}
\caption{Detailed AVUT and Video-MME (without subtitles) results for Qwen2.5-Omni-7B and 3B
in the $35\%$ and $25\%$ prefill-FLOPs tiers. EL, OR, OM, IE, CC, and CM denote Event Location,
OCR Matching, Object Matching, Information Extraction, Content Counting, and Character Matching,
respectively. Cost columns \emph{FLOPs (T)}, \emph{Ratio}, and \emph{Pre-LLM} are defined in Appendix~G.
Within each backbone and tier, the best and second-best compressors for each cost and benchmark metric are \textbf{bold} and \underline{underlined}, respectively. \emph{Full Tokens} is the uncompressed reference.}
\label{tab:avut}
\end{table*}

\begin{table*}[!t]
\centering
\renewcommand{\arraystretch}{1.0}
\setlength{\tabcolsep}{3.96pt}
{\small
\begin{tabular}{lc*{12}{r}}
\toprule
\multirow[t]{2}{*}{Method} & \multirow[t]{2}{*}{$R$ (\%)} & \multicolumn{3}{c}{Cost} & \multicolumn{9}{c}{WorldSense} \\
\cmidrule(lr){3-5}\cmidrule(lr){6-14}
 & & FLOPs (T) $\downarrow$ & Ratio $\downarrow$ & Pre-LLM $\downarrow$ & TS $\uparrow$ & CP $\uparrow$ & DL $\uparrow$ & FT $\uparrow$ & PF $\uparrow$ & GM $\uparrow$ & SP $\uparrow$ & MU $\uparrow$ & Avg. $\uparrow$ \\
\midrule
\multicolumn{14}{c}{\emph{Qwen2.5-Omni-7B}} \\
\midrule
\rowcolor{mycol2!45!white}
Full Tokens & 100 & 90.1 & 100.0\% & 100\% & 52.2 & 51.1 & 48.5 & 43.8 & 43.1 & 41.6 & 42.1 & 47.0 & 46.8 \\
\midrule
Random & 35 & \textbf{29.6} & \underline{33.4\%} & \textbf{40.0\%} & 47.1 & 46.6 & 44.8 & 41.2 & 38.6 & 39.9 & 38.6 & 44.8 & 43.2 \\ 
DyCoke (V\&A)~\citeyearpar{dycoke} & 35 & 31.2 & 35.1\% & 41.8\% & 46.1 & 46.0 & 43.2 & 40.9 & 38.6 & 39.1 & 40.0 &\underline{47.0}& 43.0 \\ 
FlashVID~\citeyearpar{flashvid} & 35 & 34.2 & 35.9\% & 47.1\% & 48.8 & 48.2 & 45.3 & 41.7 & 41.2 &\underline{42.5}&\underline{42.1}& 44.8 & 44.6 \\ 
FastV~\citeyearpar{fastv} & 35 & 34.1 & 37.8\% & 100\% & 49.0 & 47.2 & 44.8 & 41.7 & 40.1 & 40.8 & 40.5 & 45.1 & 44.1          \\ 
UniComp~\citeyearpar{unicomp} & 35 & \underline{31.0} & 34.6\% & 41.3\% & 48.4 & 48.9 & 44.5 &\underline{42.2}& 41.2 & 39.9 & 41.4 & 44.8 & 44.3 \\ 
OmniSelect~\citeyearpar{omniselect} & 35 & 35.0 & 38.8\% & 44.8\% & 47.8 & 46.4 &\underline{46.1}& 42.1 &\textbf{48.9}&\textbf{43.5}&\textbf{43.5}& 44.9 &\underline{45.6} \\
OmniZip~\citeyearpar{omnizip} & 35 & \textbf{29.6} & \textbf{32.8\%} & \underline{40.1\%} &\underline{49.4}&\textbf{50.8}& 46.0 &\textbf{44.1}& 41.9 & 40.3 & 40.2 & 46.3 & 45.3 \\ 
\rowcolor{mycol1}
\textbf{A-PACK} & 35 & \textbf{29.6} & \textbf{32.8\%} & 50.6\% &\textbf{50.8}&\underline{49.2}&\textbf{46.5}&\textbf{44.1}&\underline{43.4}&\underline{42.5}&\underline{42.1}&\textbf{47.3}&\textbf{46.1} \\
\midrule
Random & 25 & 22.1 & 25.0\% & 30.7\% & 43.9 & 43.0 & 42.7 & 39.8 & 41.2 & 37.3 & 37.4 & 45.1 & 41.6 \\ 
DyCoke (V\&A)~\citeyearpar{dycoke} & 25 & 22.5 & 25.7\% & 31.4\% & 44.7 & 41.1 & 40.0 & 36.1 & 36.3 & 39.9 & 37.0 & 44.8 & 40.3 \\ 
FlashVID~\citeyearpar{flashvid} & 25 & 24.8 & 25.2\% & 32.3\% &\underline{48.2}&\underline{46.6}& 43.6 &\underline{41.2}& 39.7 &\underline{40.3}&\underline{41.6}& 43.6 &\underline{43.5} \\
FastV~\citeyearpar{fastv} & 25 & 28.0 & 29.1\% & 100.0\% & 45.9 & 43.4 & 42.4 & 37.2 & 36.3 & 36.5 & 37.4 & 43.6 & 41.0          \\ 
UniComp~\citeyearpar{unicomp} & 25 & 23.4 & 24.8\% & \underline{30.4\%} & 47.3 & 46.0 &\underline{45.1}& 37.5 & 38.2 & 35.2 & 38.1 & 43.1 & 42.1 \\ 
OmniSelect~\citeyearpar{omniselect} & 25 & 28.7 & 31.9\% & 34.8\% & 45.8 & 40.7 & 43.7 & 35.2 &\textbf{42.4}& 36.9 & 38.6 & 40.8 & 40.9 \\ 
OmniZip~\citeyearpar{omnizip} & 25 & \underline{20.3} & \underline{22.5\%} & \textbf{28.7\%} & 46.3 & 45.0 & 43.3 & 40.9 & 38.6 & 39.5 & 39.3 &\underline{45.8}& 42.7          \\ 
\rowcolor{mycol1}
A-PACK & 25 & \textbf{19.5} & \textbf{21.6\%} & 37.5\% &\textbf{51.2}&\textbf{48.2}&\textbf{45.4}&\textbf{41.4}&\underline{41.9}&\textbf{40.8}&\textbf{42.6}&\textbf{47.3}&\textbf{45.3} \\
\midrule
\multicolumn{14}{c}{\emph{Qwen2.5-Omni-3B}} \\
\midrule
\rowcolor{mycol2!45!white}
Full Tokens & 100 & 47.3 & 100.0\% & 100\% & 51.8 & 50.8 & 45.1 & 44.9 & 42.7 & 41.2 & 44.4 & 46.1 & 46.2 \\
\midrule
Random & 35 & 16.2 & 35.0\% & 44.9\% & 49.0 & 48.9 & 41.6 & 42.0 & 40.1 &\textbf{42.5}& 40.7 & 42.9 & 43.5 \\ 
DyCoke (V\&A)~\citeyearpar{dycoke} & 35 & 16.3 & 35.0\% & 44.9\% & 48.8 & 46.9 & 40.7 & 42.2 & 39.7 &\underline{42.1}& 42.3 & 43.3 & 43.3 \\ 
FlashVID~\citeyearpar{flashvid} & 35 & 17.7 & 35.4\% & 53.9\% & 50.0 & 48.9 & 43.2 & 43.5 &\underline{41.9}& 41.2 & 40.5 & 42.1 & 44.1 \\ 
FastV~\citeyearpar{fastv} & 35 & 15.8 & 33.5\% & 100\% & 48.8 & 46.9 & 41.9 &\underline{43.8}& 41.2 & 40.3 & 39.5 & 44.6 & 43.5          \\ 
UniComp~\citeyearpar{unicomp} & 35 & 14.8 & 31.6\% & \underline{41.3\%} & 50.6 & 48.9 &\underline{44.5}& 43.5 &\textbf{42.3}& 38.6 & 42.1 & 43.6 &\underline{44.7} \\
OmniSelect~\citeyearpar{omniselect} & 35 & 16.5 & 34.8\% & 43.4\% & 49.6 & 48.9 & 42.9 & 43.0 & 39.7 & 39.5 & 42.3 &\textbf{45.8}& 44.3 \\ 
OmniZip~\citeyearpar{omnizip} & 35 & \underline{14.0} & \underline{29.7\%} & \textbf{39.9\%} &\textbf{51.8}&\textbf{52.1}& 43.9 &\textbf{44.9}& 39.7 & 41.6 &\underline{42.6}& 43.3 &\textbf{45.3} \\
\rowcolor{mycol1}
A-PACK & 35 & \textbf{13.5} & \textbf{28.6\%} & 50.6\% &\underline{50.8}&\underline{50.5}&\textbf{44.8}& 42.5 &\underline{41.9}& 40.3 &\textbf{43.3}&\underline{45.3}&\textbf{45.3} \\
\midrule
Random & 25 & 11.5 & 25.0\% & 33.8\% & 46.9 & 46.0 & 39.8 & 38.0 & 37.5 & 42.1 & 38.4 & 42.1 & 41.4 \\ 
DyCoke (V\&A)~\citeyearpar{dycoke} & 25 & 11.5 & 25.0\% & 33.8\% & 44.9 & 46.0 & 39.4 & 37.5 & 38.6 & 39.5 &\textbf{40.9}& 41.6 & 41.1 \\ 
FlashVID~\citeyearpar{flashvid} & 25 & 13.1 & 25.3\% & 38.1\% &\textbf{50.0}& 46.0 & 42.7 &\underline{41.7}&\underline{40.8}&\textbf{43.3}& 39.8 &\underline{44.6}&\underline{43.8} \\
FastV~\citeyearpar{fastv} & 25 & 12.8 & 25.0\% & 100.0\% & 47.1 & 44.7 & 42.1 &\underline{41.7}& 36.7 &\underline{42.9}& 36.5 & 41.6 & 41.9          \\ 
UniComp~\citeyearpar{unicomp} & 25 & 11.1 & 22.2\% & \underline{30.4\%} &\underline{48.2}&\underline{46.9}&\textbf{43.9}&\textbf{42.7}& 39.7 &\underline{42.9}& 36.3 & 42.6 & 43.1 \\ 
OmniSelect~\citeyearpar{omniselect} & 25 & 11.7 & 24.8\% & 32.5\% & 46.9 & 46.6 & 41.8 & 39.6 & 37.8 & 39.9 & 38.8 &\textbf{46.1}& 42.5 \\ 
OmniZip~\citeyearpar{omnizip} & 25 & \underline{9.4} & \underline{19.9\%} & \textbf{28.6\%} & 47.8 & 44.3 & 40.0 & 40.1 & 39.3 & 40.8 & 39.8 & 44.1 & 42.1          \\ 
\rowcolor{mycol1}
A-PACK & 25 & \textbf{8.7} & \textbf{18.4\%} & 37.5\% & 47.6 &\textbf{49.5}&\underline{43.0}& 41.2 &\textbf{42.7}& 41.2 &\underline{40.5}&\textbf{46.1}&\textbf{44.0} \\
\bottomrule
\end{tabular}%
}
\caption{Detailed WorldSense results across its eight domains for Qwen2.5-Omni-7B and 3B
in the $35\%$ and $25\%$ prefill-FLOPs tiers. Cost columns \emph{FLOPs (T)}, \emph{Ratio}, and \emph{Pre-LLM}
(before-LLM retention) are defined in Appendix~G. Domains: TS (Tech \& Science), CP (Culture \& Politics),
DL (Daily Life), FT (Film \& TV), PF (Performance), GM (Games), SP (Sports), MU (Music).}
\label{tab:worldsense}
\end{table*}
}

\newcommand{\TaxonomyTable}{%
\begin{table*}[!t]
\centering
\renewcommand{\arraystretch}{1.0}
{\small
\begin{tabularx}{\textwidth}{l l cc >{\raggedright\arraybackslash}X >{\raggedright\arraybackslash}X >{\raggedright\arraybackslash}X}
\toprule
\multirow[t]{2}{*}{Method} & \multirow[t]{2}{*}{Stage} & \multicolumn{2}{c}{Pruned modality} & \multirow[t]{2}{*}{Selection cue} & \multirow[t]{2}{*}{AV correspondence} & \multirow[t]{2}{*}{Budget basis} \\
\cmidrule(lr){3-4}
 & & Audio & Video & & & \\
\midrule
\rowcolor{mycol2!45!white}
Full Tokens & --- & $\times$ & $\times$ & --- & --- & --- \\
\midrule
\multicolumn{7}{c}{\emph{No audio--visual correspondence}} \\
\midrule
Random & pre-LLM & \checkmark & \checkmark & uniform & none & total-AV \\
DyCoke~\citeyearpar{dycoke} & pre-LLM & \checkmark & \checkmark & temporal & none & per-modality \\
FastV~\citeyearpar{fastv} & inner & $\times$ & \checkmark & attention & none & total-AV \\
UniComp~\citeyearpar{unicomp} & pre-LLM & $\times$ & \checkmark & visual similarity & none & total-AV \\
FlashVID~\citeyearpar{flashvid} & pre $+$ inner & $\times$ & \checkmark & hierarchical & none & video-only \\
\midrule
\multicolumn{7}{c}{\emph{Static, pointwise audio--visual correspondence}} \\
\midrule
OmniSelect~\citeyearpar{omniselect} & pre-LLM & \checkmark & \checkmark & A: attn.; V: cosine & static pointwise & per-modality \\
OmniZip~\citeyearpar{omnizip} & pre-LLM & \checkmark & \checkmark & audio saliency & static pointwise & per-modality \\
\multicolumn{7}{c}{\emph{Local-window audio--visual correspondence}} \\
\midrule
\rowcolor{mycol1}
\textbf{A-PACK (ours)} & pre $+$ inner & $\times$\,/\,\checkmark & \checkmark & local CKA; query attn. & local-window CKA & total-AV \\
\bottomrule
\end{tabularx}%
}
\caption{Taxonomy of the evaluated training-free methods.}
\label{tab:taxonomy}
\par
\refstepcounter{section}\label{sec:reproducibility}
{\raggedright\noindent\textbf{\thesection\quad Method Taxonomy and Qualitative Analysis}\par}
\begin{minipage}[t]{0.485\textwidth}
\refstepcounter{subsection}\label{sec:taxonomy}
\noindent\textbf{\thesubsection\quad Method Taxonomy}\par
Table~\ref{tab:taxonomy} compares the evaluated methods across four~decisions: when compression occurs, which modality is pruned, how audio--visual correspondence is modeled, and how the token budget is defined. The selection-cue column specifies the score used to implement these choices. These axes separate methods that may appear similar from their final retention alone. In particular, a pre-LLM method must decide before query-conditioned cross-modal interactions are fully formed, whereas an inner-LLM method can use contextualized attention but retains the early-layer computation.

The modality columns clarify the design consequence of this timing. Visual-only methods avoid direct audio loss but cannot reduce audio tokens, while joint pre-LLM methods can compress both streams before their relevance is resolved. A-PACK instead marks audio as unpruned before the LLM and prunable after fusion. This is not a permanent preference for audio; it delays the decision until query attention can assess audio and visual evidence in a shared context.

The correspondence column further distinguishes the information used for visual allocation. OmniSelect and OmniZip use static pointwise audio--visual cues, which can associate semantically related events even when they occur at different times. A-PACK uses local-window CKA to favor co-timed representational changes, then handles within-video redundancy separately. Thus, cross-modal correspondence determines where visual capacity is allocated, while anchor-based selection determines which visual tokens are retained within that allocation.

The budget basis specifies which tokens share the compression target. A total-AV budget limits the combined number of audio and visual tokens, a per-modality budget assigns separate targets to the two streams, and a video-only budget leaves audio outside the target. Consequently, the same nominal retention value can represent different sequence lengths and computational costs. We therefore compare methods at matched prefill-FLOPs tiers and report the realized cost together with token retention. Under this common cost measure, A-PACK's efficiency comes from early visual reduction followed by later multimodal pruning, rather than from assigning an unusually small budget to one modality. Code will be released upon acceptance.
\end{minipage}\hfill
\begin{minipage}[t]{0.485\textwidth}
\refstepcounter{subsection}\label{sec:qualitative-analysis}
\noindent\textbf{\thesubsection\quad Qualitative Analysis}\par
Figures~\ref{fig:qualitative}--\ref{fig:qualitative-failure} use a three-row layout. The first row gives the sample, question, and reference answer. The second presents OmniZip's prediction, key-frame visual-pruning mask, and audio-preservation curve over time; the third presents the same information for A-PACK.

The mask indicates whether query-relevant visual evidence survives at the key frame, while the curve shows how much audio is retained around the same interval. Reading them together reveals whether both modalities provide the evidence needed for the displayed prediction.

In Figure~\ref{fig:qualitative}, the phrase ``This man retired'' identifies the relevant person and scene. OmniZip removes audio evidence near this event and prunes the associated visual content, leading to ``Nobody.'' A-PACK preserves the spoken cue before fusion and retains the locally corresponding old-man frame, providing both pieces of evidence required for the correct response.

Figure~\ref{fig:qualitative-second} asks for the interval containing crisp clapping. OmniZip's early audio removal weakens the temporal cue and its visual pruning loses relevant key-frame evidence. A-PACK retains the audio guidance and corresponding visual region, allowing the model to select the relevant interval.

Figure~\ref{fig:qualitative-failure} shows the limitation. Although A-PACK preserves pre-LLM audio and aligned visual context, it misses the second-level location of a laugh. Its local window retains the broader event neighborhood but does not explicitly protect the exact acoustic transition. Fine temporal evidence may also be removed by progressive inner-LLM pruning after fusion.

The two successes illustrate complementary roles for deferred audio pruning and Gated CKA, whereas the failure separates coarse event preservation from exact temporal localization. Retaining the relevant interval and resolving an exact timestamp are distinct requirements: the latter also requires protecting brief transitions after fusion. This qualitative evidence indicates whether an error follows early evidence removal or insufficient temporal precision after evidence is retained. Future work should combine query-adaptive temporal granularity with layer-wise retention that protects short audio events when precise timing is required.\end{minipage}
\end{table*}
}

\section{Efficiency and Cost Measurement}
\label{sec:cost-evaluation}
Accuracy is evaluated with the same pipeline and decoding protocol across all methods and compression settings.
Our efficiency protocol follows the analytic, token-count accounting of recent holistic token-merging
work \citep{holitom}, adapted to an omni-modal (audio\,+\,video) model. We report prefilling FLOPs and
the FLOPs ratio, the before-LLM retained-token ratio, time-to-first-token (TTFT) with a module-level
breakdown, decoding throughput, end-to-end latency, and peak GPU memory. The before-LLM
retained-token ratio is the proportion of original audio--visual tokens that enter the LLM, and it directly
determines the prefill sequence length.
Unless otherwise noted, we measure CUDA-synchronized end-to-end wall-clock latency on Qwen2.5-Omni-7B
using bfloat16, FlashAttention-2, and greedy decoding on an NVIDIA A6000 GPU at the $35\%$ and $25\%$
prefill-FLOPs tiers.

\subsection{FLOPs Accounting}
\paragraph{Prefill FLOPs.} We estimate compute analytically from token counts, following the
holistic token-merging accounting of \citet{holitom} and adapting it to an omni-modal model. One
transformer layer at token length $n$ costs
\begin{equation}
\phi(n) \;=\; \underbrace{4 n d^{2}}_{\text{Q,K,V,O proj.}}
        \;+\; \underbrace{2 n^{2} d}_{\text{attention}}
        \;+\; \underbrace{2 n d m}_{\text{FFN}},
\label{eq:perlayer}
\end{equation}
with hidden size $d$ and FFN width $m$. The three terms are the four attention projections, the
quadratic attention, and the feed-forward block. Prefill FLOPs sum this over the \emph{actual}
per-layer token schedule $\{n_i\}_{i=1}^{T}$. This accounting assigns each method the compute
incurred at every layer before and after its pruning events:
\begin{equation}
F_{\mathrm{pre}} \;=\; \sum_{i=1}^{T} \phi(n_i),
\qquad
F_{\mathrm{pre}}^{\mathrm{full}} \;=\; T\,\phi(n_0),
\label{eq:prefill}
\end{equation}
where $n_i$ is the length entering layer $i$ and $n_0$ is the uncompressed input length. The reported
\emph{FLOPs Ratio} is prefill-only,
\begin{equation}
\rho_{\mathrm{FLOPs}} \;=\; \frac{F_{\mathrm{pre}}}{F_{\mathrm{pre}}^{\mathrm{full}}}
\;=\; \frac{\sum_{i=1}^{T}\phi(n_i)}{T\,\phi(n_0)},
\label{eq:ratio}
\end{equation}
and the \emph{Prefilling FLOPs} column reports the numerator $F_{\mathrm{pre}}$ in TFLOPs. The
constants are $T{=}28$, $d{=}3584$, $m{=}18944$ for the Qwen2.5-Omni-7B thinker and $T{=}36$,
$d{=}2048$, $m{=}11008$ for 3B. Decoding is fixed at $R{=}100$ generated tokens and is excluded from
the prefill ratio.

\paragraph{Per-layer token schedule.} Equation~\eqref{eq:prefill} places every method on one basis
through its schedule $\{n_i\}$. Input-level pruners carry a single post-prune length across all $T$
layers, so $n_i$ is constant. A method that prunes at an intermediate layer keeps the full length
until that layer, so FastV holds $n_i{=}n_0$ up to layer~$3$ and FlashVID up to layer~$20$. A-PACK
prunes progressively from $L_{\mathrm{mid}}$, so its $\{n_i\}$ steps down across the upper layers. Each
schedule is read from the recorded per-sample token logs, so $\rho_{\mathrm{FLOPs}}$ reflects measured
retention rather than the nominal knob. 

\subsection{Runtime and Memory Measurement}
\paragraph{Prefill / TTFT and its breakdown.} Under greedy decoding, TTFT is the prefill time
until the first output token becomes available. We measure wall time from entering \texttt{generate()} to
the first \texttt{lm\_head} forward, when the first-token logits are produced. Per-module hooks separate
the vision tower, the LLM backbone, and other processing. Other includes the pre-LLM compression pipeline
and its preprocessing overhead. The LLM backbone includes A-PACK's inner-LLM pruning, so its reduction
captures the benefit of progressively shortening the later-layer sequence.

\paragraph{Throughput, latency, memory.} End-to-end latency is the total \texttt{generate()}
wall time. Decoding time is the remaining time after TTFT, and throughput is the number of generated tokens
per decoding second. The pre-LLM stage reduces the cost paid before decoding begins, whereas inner-LLM
pruning reduces the cached sequence used by every subsequent decoding step. Peak GPU memory complements
FLOPs by measuring the runtime memory demand of the resulting cache.

\ExpandedBenchmarkTables
\TaxonomyTable
\FloatBarrier

\setcounter{figure}{7}
\begin{figure*}[p]
\centering
\includegraphics[width=0.78\textwidth,height=0.88\textheight,keepaspectratio]{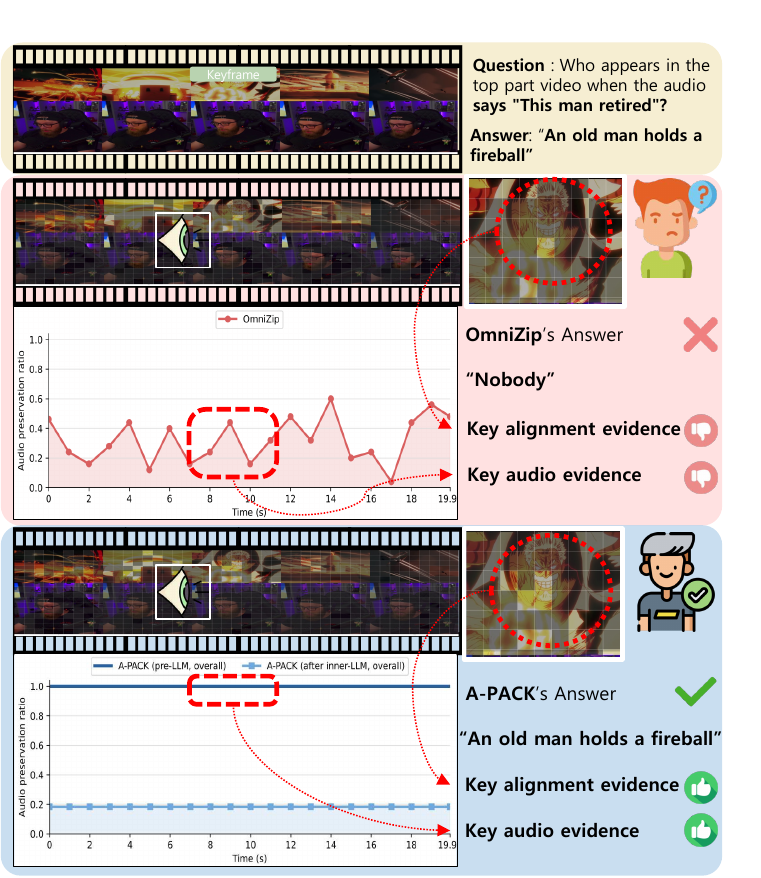}
\caption{Qualitative success on AVUT. OmniZip removes audio evidence that identifies the relevant visual moment, whereas A-PACK preserves the spoken cue before the LLM and retains its locally aligned visual evidence, yielding the correct answer.}
\label{fig:qualitative}
\end{figure*}

\begin{figure*}[p]
\centering
\includegraphics[width=0.78\textwidth,height=0.88\textheight,keepaspectratio]{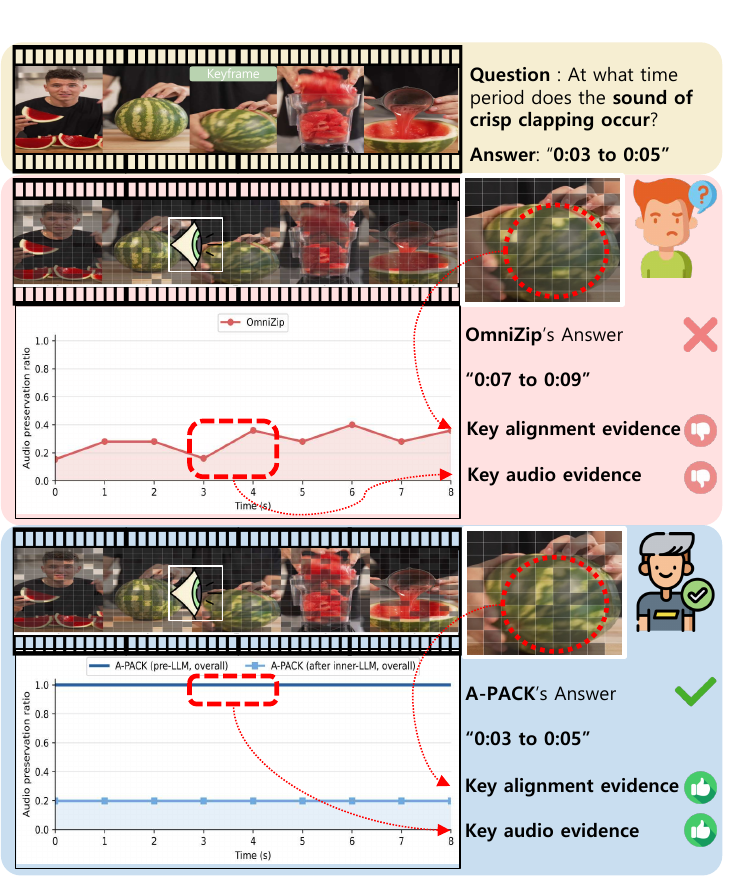}
\caption{Qualitative success on AVUT involving temporal audio evidence. Preserved audio and locally aligned frame allocation retain the evidence needed for the answer, which early audio pruning can lose.}
\label{fig:qualitative-second}
\end{figure*}

\begin{figure*}[p]
\centering
\includegraphics[width=0.78\textwidth,height=0.98\textheight,keepaspectratio]{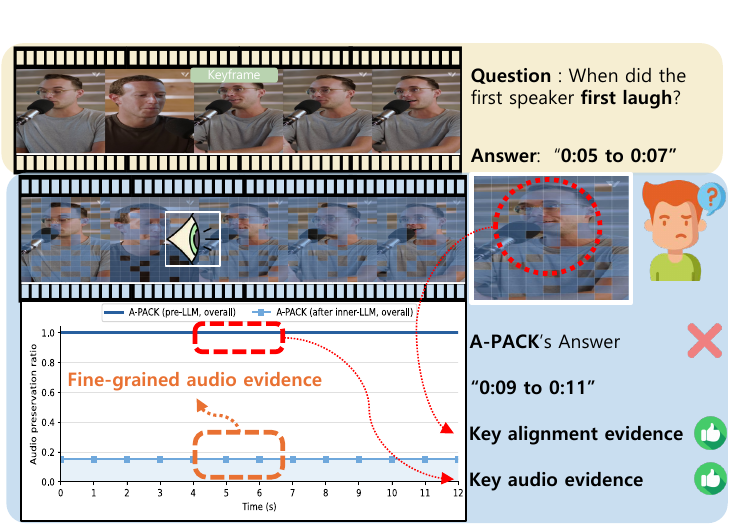}
\caption{Failure case on AVUT. A-PACK misses second-level temporal localization of the speaker's laugh despite retaining relevant audio and aligned visual evidence, motivating temporally finer allocation and inner-LLM pruning.}
\label{fig:qualitative-failure}
\end{figure*}

\end{document}